%% file: main.tex
\documentclass[10pt,twocolumn,letterpaper]{article}

\usepackage[pagenumbers]{cvpr} 

\input{preamble}
\withappendixfalse

\definecolor{cvprblue}{rgb}{0.21,0.49,0.74}
\usepackage[pagebackref,breaklinks,colorlinks,allcolors=cvprblue]{hyperref}

\def\confName{CVPR}
\def\confYear{2027}

\title{FCEdit: Frequency-Calibrated Endpoint Correction for Source-Consistent Inversion-Free Flow Editing}

\author{
Soyun Lee$^{1,2}$\quad
Chaewon Park$^{1}$\quad
Sehwa Lee$^{1}$\quad
Tae-Kyun Kim$^{2^{*}}$\quad
Kibeom Hong$^{1^{*}}$\\[0.5em]
{ $^{1}$Sookmyung Women's University}\quad
{ $^{2}$KAIST}\\[0.3em]
{\small\tt \{synuolee, kimtaekyun\}@kaist.ac.kr,\ \{sea9392, 1ee3flowers, kb.hong\}@sookmyung.ac.kr}
}

\begin{document}
\maketitle
\begin{NoHyper}\renewcommand{\thefootnote}{*}\footnotetext{ Corresponding authors.}\end{NoHyper}
\input{sec/0_abstract}    
\input{sec/1_intro}

\input{sec/2_related}
\input{sec/3_method}
\input{sec/4_exp}

\input{sec/5_conclusion}
{
    \small
    \bibliographystyle{ieeenat_fullname}
    \bibliography{main}
}

\ifwithappendix
\clearpage
\appendix
\input{sec/X_appendix}
\fi


\end{document}

%% file: preamble.tex
\usepackage{algorithm}
\usepackage{algorithmic}
\newcommand{\cmark}{\ensuremath{\surd}}
\newcommand{\xmark}{\ensuremath{\times}}
\usepackage{placeins}
\newif\ifwithappendix \withappendixtrue
\newcommand{\suppref}[1]{\ifwithappendix Appendix~\ref{#1}\else the supplementary material\fi}

\definecolor{revpink}{RGB}{231,41,138}

\usepackage{tikz}
\usetikzlibrary{calc}
\definecolor{qualpurple}{RGB}{166,40,157}
\newlength{\qualw}
\newcommand{\qualimg}[1]{\includegraphics[width=\qualw]{#1}}
\newcommand{\qualcirc}[5]{%
  \begin{tikzpicture}[inner sep=0pt,outer sep=0pt,baseline=0pt]
    \node[anchor=south west,inner sep=0pt] (qualimgnode) at (0,0) {\includegraphics[width=\qualw]{#1}};
    \draw[red,dash pattern=on 1.4pt off 1pt,line width=1.5pt]
      ($(qualimgnode.south west)+(#2\qualw,#3\qualw)$) ellipse [x radius=#4\qualw,y radius=#5\qualw];
  \end{tikzpicture}%
}

%% file: sec/0_abstract.tex
\begin{abstract}

Inversion-free text-guided image editing with pretrained rectified-flow models modifies real images without costly inversion or image-specific optimization. However, errors accumulated along the sampling trajectory cause preserved content to drift from the original image. 
Existing remedies mitigate this drift with a constant or time-scheduled scalar correction, treating every frequency component as equally reliable, which locks editing into a rigid trade-off between source fidelity and editability.
We propose FCEdit, an inversion-free editing framework that calibrates this endpoint correction across time and frequency. 
A closed-form Wiener gain, built from a spectral prior on the latent and an uncertainty model of the learned flow, determines how far each frequency component of the correction should be trusted. This gain recovers prior scalar schedules as frequency-flat special cases. 
An energy-guided soft decomposition further allocates editing and correction signals without external masks or extra network evaluations.
On PIE-Bench, FCEdit substantially improves source consistency while maintaining competitive editability, reducing background LPIPS by 12\% relative to the strongest baseline. It is also the most preferred method in a human study.
\end{abstract}

%% file: sec/1_intro.tex
\section{Introduction}
\label{sec:intro}
\begin{figure}[t]
\centering
\includegraphics[width=\columnwidth]{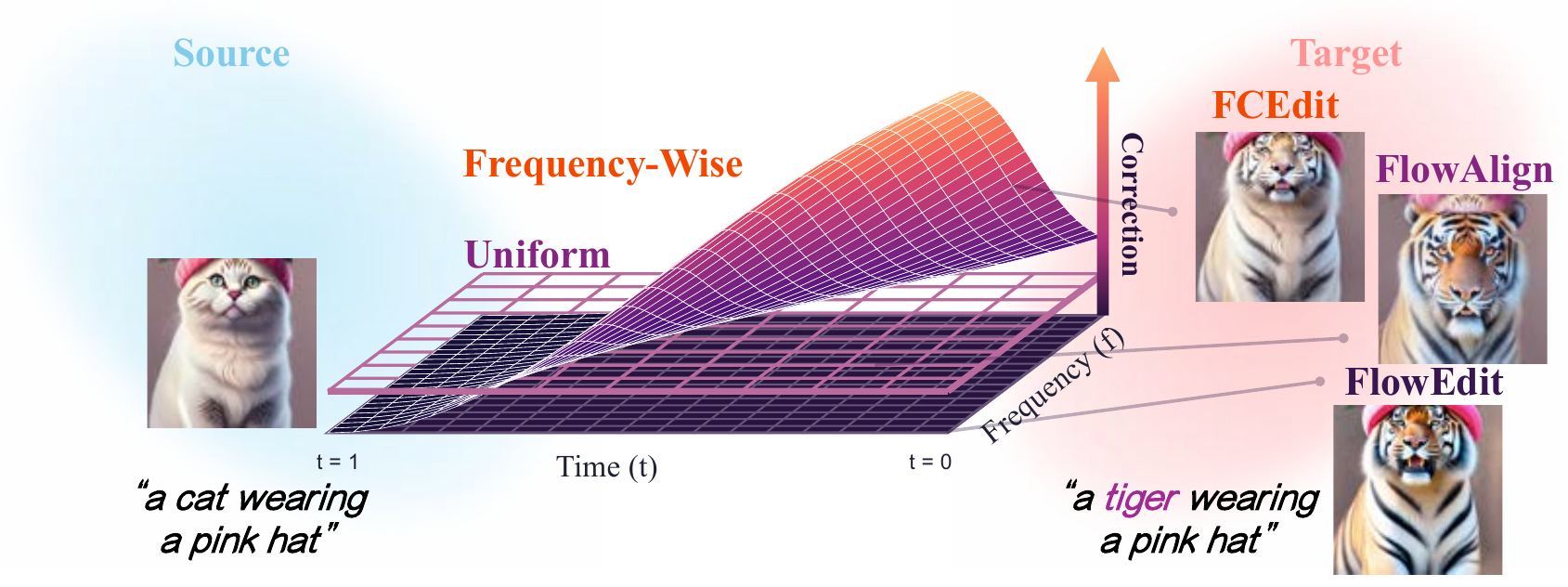}
\caption{\textbf{Frequency-wise endpoint correction for source-consistent image editing.}
In inversion-free editing process, FlowEdit~\cite{flow_edit} enable semantic changes but can drift far from the source along the editing trajectory.
FlowAlign~\cite{flow_align} uses endpoint correction to mitigate this drift, but its fixed strength can also suppress intended edits.
FCEdit instead calibrates endpoint correction across timesteps and frequencies using the Wiener gain, suppressing unreliable responses while retaining source-restoring information.
FCEdit better achieves intended semantic change (\textit{cat-to-tiger}) while preserving source details (\textit{pink hat, background}).}
\label{teaser}

\end{figure}

Text-guided image editing{~\cite{mengsdedit,brooks2023instructpix2pix,null_text_inversion,flow_edit,bfl2025kontext}} aims to realize textual instruction while preserving contents unrelated to the requested change. 
On U-Net-based diffusion models{~\cite{ho2020ddpm,songdenoising,rombach2022ldm}}, training-free editors pursue this balance by intervening in the model's internal computation through attention control~\cite{hertzprompt,masactrl,pix2pix_zero}, feature injection~\cite{plug_and_play}, or explicit mask guidance~\cite{diff_edit,blended_diffusion}.
Such interventions tie editing to a particular architecture, an explicit localization step, or image-specific optimization. As generation paradigm shifted to rectified-flow transformers~\cite{peebles2023dit,SD3,SD35_medium,flux1_dev}, a complementary line of work edits the sampling trajectory
itself through the model's velocity predictions. 

In flow-based editing~\cite{lipman2023flow,liuflow}, a common route to source fidelity is to invert the source image into a noise latent and then regenerate it under the target prompt~\cite{null_text_inversion,wallace2023edict,huberman2024editfriendly,Rout2024SemanticII,wang2025taming,dengfireflow}. However, the inversion process adds a costly pre-editing stage, accumulates trajectory error, and does not itself guarantee faithful editing. FlowEdit~\cite{flow_edit} removes this stage by directly integrating the difference between target- and source-conditioned velocities. At each timestep, it mixes the source latent with fresh Gaussian noise to obtain a valid source marginal, eliminating the need for an image-specific inversion trajectory.

Unfortunately, marginal validity alone does not guarantee trajectory consistency. Source states are independently resampled at every timestep, and the guided velocities acting on them are imperfect predictions, so errors accumulate along the trajectory and preserved content drifts from the source. To mitigate this drift, recent inversion-free methods introduce different forms of source control. FlowAlign~\cite{flow_align} feeds a source-directed endpoint correction back into the editing trajectory with a fixed strength, while DRFS~\cite{drfs} applies a time-dependent trajectory shift. FIA-Edit~\cite{fia_edit} combines frequency interaction with feature injection, and ChordEdit~\cite{chord_edit} regularizes the editing transport field. In all of these, some correction toward the source is fed back into the editing trajectory at a chosen strength.

Notably, this strength is either a single global coefficient or a time-dependent scalar schedule, so every frequency component at a given timestep is corrected with the same weight. 
Yet the correction signal itself does not warrant this uniformity. It mixes source-restoring information with endpoint-estimation error and edit-related deviations, whose relative strength varies across timesteps and frequencies. A scalar strength therefore cannot suppress the unreliable components without also suppressing the reliable ones. The same tension arises in space, where a uniformly applied correction pulls edited regions back toward the source, and explicit masks avoid this only by adding an external localization step.

Our key observation is that deciding how far to trust the correction is an estimation problem rather than a design choice. The correction signal is a noisy observation of the source-restoring component it is meant to supply. Based on this insight, we propose \textbf{FCEdit}, an inversion-free editor that calibrates endpoint correction across timesteps, frequencies, and spatial locations.  

Specifically, Frequency-Wise Wiener Correction (FWC) models the signal and error powers of the correction with a structural prior on the source spectrum and an uncertainty model of the learned flow, from which the correction weight follows in closed form as a Wiener gain rather than a hand-designed schedule. 
Soft Gate Decomposition (SGD) further complements it with an energy-guided soft decomposition that treats target-directed editing and source-consistency correction as competing local proposals and allocates responsibility between them from their local energies. Both energies derive from signals already computed during sampling, so the allocation requires no external mask, auxiliary network, or per-image optimization.

We extensively validate the effectiveness of FCEdit on various recent inversion-free flow models, including FlowEdit~\cite{flow_edit}, FlowAlign~\cite{flow_align}, FIA-Edit~\cite{fia_edit}, DRFS~\cite{drfs}, and inversion-based RF-Inversion~\cite{Rout2024SemanticII}. On PIE-Bench~\cite{pie_bench}, FCEdit substantially improves source preservation while maintaining competitive editability, staying on top across other methods. For instance, we demonstrate that FCEdit reduce background LPIPS by 12\% over the strongest baseline, while achieving better edits.  Ablations attribute the largest share of this improvement to frequency calibration, with structure distance more than doubling under a frequency-flat correction.

In summary, our contribution is given as follows:
\begin{itemize}
    \item We propose \textbf{FCEdit}, an inversion-free editor that recasts endpoint correction as a reliability-estimation problem and solves it with a closed-form time--frequency Wiener gain, of which the constant and time-only schedules of prior work are frequency-flat special cases.
    \item We introduce an energy-guided soft decomposition that spatially allocates editing and correction from signals already computed during sampling, without external masks, auxiliary networks, or extra neural-function evaluations.
    \item On PIE-Bench~\cite{pie_bench}, FCEdit achieves the best structural consistency, perceptual background fidelity, and whole-image CLIP similarity, and is the most preferred method in human study.
\end{itemize}

%% file: sec/2_related.tex
\section{Related Work}
\subsection{Inversion-Free Text-Guided Image Editing}
Inversion-based editors~\citep{wang2025taming, dengfireflow, ft_edit, jiao2026unieditflow, yangdirectedit} first recover a source-specific noise state through an explicit inversion process and then edit along the resulting trajectory. They improve inversion accuracy or reuse features along the inverted trajectory. Another line controls editing through internal attention in DiT backbones~\citep{avrahami2025stableflow, kv_edit, exploring_mmdit}.

Inversion-free editors~\citep{xu2024infedit, mengsdedit, su2022dual, flow_edit, split_flow, flow_align, fia_edit, drfs} instead avoid explicitly recovering the source noise state. For diffusion models, InfEdit~\citep{xu2024infedit} removes explicit inversion through a consistency-model sampling scheme combined with unified attention control. SDEdit~\citep{mengsdedit} renoises the source to an
intermediate timestep and denoises under the target prompt, trading fidelity for simplicity. DDIB~\citep{su2022dual} chains two independent diffusion bridges. 
Most relevant to our work, FlowEdit~\citep{flow_edit} exploits the closed-form interpolation of rectified flow. At every step it couples a source-side and a target-side noisy latent through shared, freshly sampled noise and integrates only their velocity difference. SplitFlow~\citep{split_flow} extends this direct scheme to complex
instructions by decomposing the target prompt into sub-prompts and aggregating their sub-flows. 

Because the source marginal in FlowEdit is independently resampled
across timesteps, its approximate editing trajectory may be unstable and
can lose source consistency. Subsequent inversion-free methods introduce
different forms of trajectory or feature control. FlowAlign~\citep{flow_align} derives a terminal source-similarity regularization
from an optimal-control formulation and applies the resulting endpoint
correction with a globally shared source-consistency scale. FIA-Edit~\citep{fia_edit} exchanges frequency components between source and
target features within self-attention and injects source-side features
into the target branch. DRFS~\citep{drfs} introduces a time-dependent
shift that moves noisy latents toward the target trajectory; disabling the shift connects the formulation to score-distillation sampling~\citep{hertz2023dds},
whereas a particular linear shift schedule recovers FlowEdit.

Viewed together, this line strengthens the source consistency by choosing how strongly the pull toward the source acts, or by choosing which features carry it. How that pull should be distributed across frequency has not been addressed deeply. FIA-Edit comes closest, yet it uses frequency as a heuristic feature pathway inside attention rather than as a principle that determines the correction itself. Spatial control, in turn, is consistently delegated to attention maps or auxiliary modules. FCEdit fills both gaps. It keeps the endpoint comparison of FlowAlign as the correction direction, but derives the gain on that correction in closed form from a power-law spectral prior verified in the latent, with the constant and time-only schedules of prior work recovered as frequency-flat special cases. Its spatial allocation is read directly from the two update streams, so no external localizer is required.

\subsection{Frequency Priors and Wiener Estimation}
The statistics of natural images are strongly ordered in frequency. Their power spectra follow an approximate $1/f^2$
law~\citep{Mandelbrot1967SomeNW,wiener1949extrapolation,Schaaf1996ModellingTP,Simoncelli2001NaturalIS,torralba2003statistics}, so low frequencies carry most of the signal energy. Classical estimation theory turns such a prior into an optimal restoration rule. Given the signal and noise spectra, the linear estimator that minimizes mean squared error is the Wiener filter~\citep{Haykin1986AdaptiveFT, Goldstein1998AMR}. Its gain reflects the relative signal and noise powers at each frequency, assigning larger weights to signal-dominated components and attenuating noise-dominated ones.
Unlike a fixed low-pass filter, this estimator determines attenuation from the relative signal and error powers rather than from frequency alone, allowing the same frequency component to receive different weights as its uncertainty changes.
Diffusion and flow models interact with this structure in a characteristic way. Their corruption is white while the signal is not, so generation recovers low frequencies before high ones and proceeds coarse to fine~\citep{rissanen2023generative, Hoogeboom2022BlurringDM, falck2025fourier}.

We apply this view to inversion-free editing, where the correction faces exactly this estimation problem. The correction estimates clean source content from a latent whose noise level is known, so how far to trust it at each timestep and frequency is dictated by the prior and the noise rather than by design choices. Where prior editing work touches frequency, it does so heuristically, through fixed band splits and feature-level injection~\citep{fia_edit}, frequency-selective score distillation~\citep{ren2025fds}, or hand-tuned time schedules~\citep{flow_align,drfs}. We instead verify a power-law prior in the VAE latent where the correction operates and obtain the gain in closed form. Classical Wiener filtering assumes the spectra are known, and our contribution is to instantiate them for a pretrained flow with a structural source prior and a calibrated uncertainty model.

%% file: sec/3_method.tex
\section{Method}
\label{sec:method}

\begin{figure*}[t]
    \centering
    \includegraphics[width=0.95\textwidth]{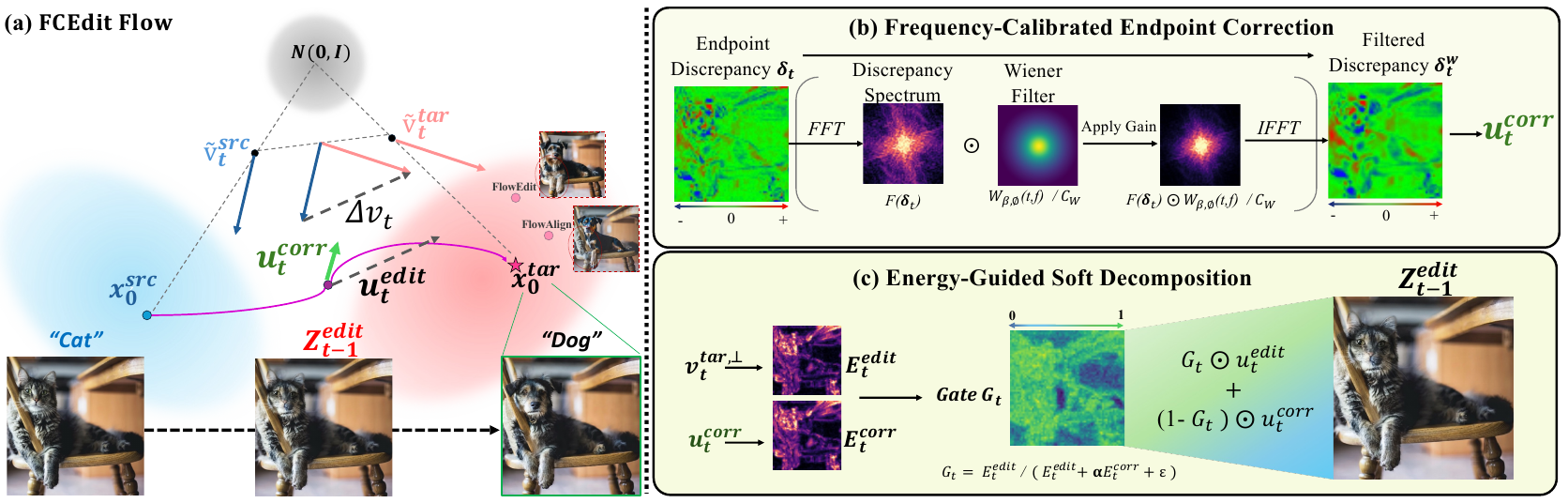}
    \caption{ \textbf{Overview of FCEdit.}
    (a) At each active sampling step $t$, FCEdit constructs source and target states and obtains their velocities under shared random noise. The velocity difference defines the editing direction, while the two clean-endpoint estimates define the endpoint discrepancy $\delta_t$.
    (b) FCEdit transforms this endpoint discrepancy into the frequency domain and applies the normalized time--frequency Wiener gain
    $W_{\beta,\phi}(t,f)/C_W$. The filtered correction proposal
    $u_t^{\mathrm{corr}}=\alpha\bar{\delta}_t^{\,w}$
    suppresses unreliable high-frequency responses while restoring
    source-consistent structure and background content that drift during
    editing.
    (c) Source-orthogonal editing evidence and correction evidence determine
    the mask-free spatial gate $G_t$, which preserves strong target-directed
    updates at edited locations while applying the source-restoring
    correction where editing evidence is weaker.
    The two proposals are softly combined to produce a target-aligned edit while reducing unintended source drift, without additional neural-function
    evaluations.
    }
    \label{fig:fig_arch}
\end{figure*}

FCEdit builds on the endpoint-based source-consistency correction introduced by FlowAlign~\cite{flow_align}. Our contribution lies in determining how strongly each time--frequency components of the correction should be trusted, and how the editing and correction proposals should be balanced across spatial locations.
We first derive the ideal frequency-wise reliability of the correction from a component-wise Wiener estimator (Sec.~\ref{sec:wiener}), adapting its unknown uncertainty power to the non-ideal prediction errors of a learned flow model. We then spatially allocate the target-directed editing and source-consistency correction proposals through an energy-guided soft decomposition (Sec.~\ref{sec:decomposition}), improving source consistency while retaining editability.

\subsection{Preliminaries}
\label{sec:preliminary}
A rectified flow model~\cite{lipman2023flow,liuflow} transports noise at $t=1$ to data at $t=0$ through
\begin{equation}
    \frac{dZ_t}{dt}
    =
    v_\theta(Z_t,t,c),
    \qquad
    Z_t
    =
    (1-t)X_0+t\epsilon,
    \quad
    \epsilon\sim\mathcal{N}(0,I),
    \label{eq:rf}
\end{equation}
where $c$ denotes the conditioning input. Under this interpolant, the ideal velocity $v_t^\star=\epsilon-X_0$ inverts to $X_0=Z_t-t\,v_t^\star$; substituting the learned $v_\theta$ yields the clean-endpoint estimate $\widehat{x}_0(Z_t,t,c)=Z_t-t\,v_\theta(Z_t,t,c)$.

Given a source latent $x_{\mathrm{src}}$, source prompt $c_{\mathrm{src}}$, and target prompt $c_{\mathrm{tar}}$, we follow the direct editing construction of FlowEdit~\cite{flow_edit}. At each active timestep $t_i$, we construct
\begin{equation}
\begin{aligned}
    z_i^{\mathrm{src}}
    &=
    (1-t_i)x_{\mathrm{src}}+t_i\epsilon_i,\\
    z_i^{\mathrm{tar}}
    &=
    Z_i^{\mathrm{edit}}
    +z_i^{\mathrm{src}}
    -x_{\mathrm{src}},
\end{aligned}
\qquad
\epsilon_i\sim\mathcal{N}(0,I).
\label{eq:flowedit_states}
\end{equation}
The source and target branches share the same $\epsilon_i$ within
each timestep, while the noise is independently resampled across
timesteps. The target state therefore preserves the current editing
displacement:
\begin{equation}
    z_i^{\mathrm{tar}}-z_i^{\mathrm{src}}
    =
    Z_i^{\mathrm{edit}}-x_{\mathrm{src}}.
    \label{eq:shared_displacement}
\end{equation}
For $b\in\{\mathrm{src},\mathrm{tar}\}$, the unconditional velocity
$v_i^{b,u}$, the conditional velocity $v_i^{b,c}$, and the CFG-guided
velocity~\cite{Ho2022ClassifierFreeDG} $\widetilde v_i^b$ with guidance scale $\eta_b$ are
\begin{equation}
\begin{aligned}
    v_i^{b,u}
    &=
    v_\theta(z_i^b,t_i,\varnothing),
    \qquad
    v_i^{b,c}
    =
    v_\theta(z_i^b,t_i,c_b),\\
    \widetilde v_i^b
    &=
    v_i^{b,u}
    +
    \eta_b
    \left(
        v_i^{b,c}-v_i^{b,u}
    \right).
\end{aligned}
\label{eq:cfg_velocities}
\end{equation}
With
$\Delta t_i=t_{i-1}-t_i<0$, the target-directed editing proposal is
\begin{equation}
    u_i^{\mathrm{edit}}
    =
    \Delta t_i
    \left(
        \widetilde v_i^{\mathrm{tar}}
        -
        \widetilde v_i^{\mathrm{src}}
    \right).
    \label{eq:edit_proposal}
\end{equation}
Although Eq.~\eqref{eq:edit_proposal} supplies a target-directed
editing displacement, it does not determine which components should
be modified and which should remain consistent with the source. To address this, we construct a frequency-shaped source-consistency correction (Sec.~\ref{sec:wiener}) and then spatially allocate the editing and correction proposals (Sec.~\ref{sec:decomposition}).

\subsection{Frequency-Wise Wiener Correction (FWC)}
\label{sec:wiener}
Building on FlowAlign's source-consistency regularization based on endpoint comparison~\cite{flow_align}, we define a source--target endpoint discrepancy adapted to the guided direct-editing trajectory. Using the raw source-conditional velocity $v_i^{\mathrm{src},c}$ from Eq.~\eqref{eq:cfg_velocities}, the discrepancy is defined as follows:
\begin{equation}
    \delta_i
    =
    \left(
        z_i^{\mathrm{src}}
        -
        t_i v_i^{\mathrm{src},c}
    \right)
    -
    \left(
        z_i^{\mathrm{tar}}
        -
        t_i\widetilde v_i^{\mathrm{tar}}
    \right).
    \label{eq:endpoint_discrepancy}
\end{equation}
The raw source-conditional endpoint provides a non-extrapolated source reference, whereas the guided target endpoint corresponds to the editing trajectory actually being followed. Using Eq.~\eqref{eq:shared_displacement}, the discrepancy can equivalently
be written as
\begin{equation}
    \delta_i
    =
    x_{\mathrm{src}}
    -
    Z_i^{\mathrm{edit}}
    +
    t_i
    \left(
        \widetilde v_i^{\mathrm{tar}}
        -
        v_i^{\mathrm{src},c}
    \right).
    \label{eq:discrepancy_decomposition}
\end{equation}
Hence, $\delta_i$ captures both accumulated source deviation and the endpoint mismatch between the source-conditioned reference and the guided target trajectory. Rather than applying this discrepancy uniformly, we estimate the reliability of its source-restoring components across timesteps and frequencies.

\paragraph{Wiener formulation of correction reliability.}
The discrepancy $\delta_i$ contains a model-error-free endpoint
discrepancy together with endpoint-prediction error, and their relative
powers need not be uniform across frequency. We therefore write
\begin{equation}
    \widehat\delta_i(f)
    =
    r_i(f)+e_i(f),
\end{equation}
where $r_i(f)$ denotes the Fourier coefficient of the model-error-free
source--target endpoint discrepancy and $e_i(f)$ denotes the residual
induced by branchwise endpoint-prediction errors.
The detailed algebraic derivation of this decomposition from the two endpoint estimates can be found in \suppref{app:discrepancy_derivation}.
For a linear estimator $\widehat r_i(f)=W(t_i,f)\widehat\delta_i(f)$, assume that $r_i(f)$ and $e_i(f)$ are zero-mean and uncorrelated, with powers $P_r(t_i,f)$ and $P_e(t_i,f)$. Minimizing $\mathbb{E}\left[\left|W(t_i,f)\widehat\delta_i(f)-r_i(f)\right|^2\right]$
yields the componentwise Wiener gain, the scalar linear minimum mean squared
error (MMSE) solution (see \suppref{app:wiener_derivation})
\begin{equation}
    W^\star(t,f)
    =
    \frac{P_r(t,f)}{P_r(t,f)+P_e(t,f)}.
    \label{eq:oracle_wiener}
\end{equation}
For vector-valued latent coefficients, the general linear-MMSE estimator would involve dense cross-channel covariance matrices. We instead adopt a frequency-diagonal, channel-isotropic proxy covariance model and broadcast the same scalar gain across latent channels. Under this proxy model, Eq.~\eqref{eq:oracle_wiener} is the exact componentwise linear-MMSE solution.

\paragraph{Rectified-flow proxy powers.}
Under the ideal interpolation in Eq.~\eqref{eq:rf}, the data and noise amplitudes scale with $(1-t)$ and $t$, respectively. We use this schedule to define the proxy powers
\begin{equation}
    P_r^{\mathrm{ideal}}(t,f)
    =
    (1-t)^2S_\beta(f),
    \qquad
    P_e^{\mathrm{ideal}}(t,f)
    =
    t^2,
    \label{eq:ideal_powers}
\end{equation}
where $S_\beta(f)=\bigl(r_{\mathrm{ref}}/r(f)\bigr)^{\beta}$ and $r(f)$ is the Nyquist-normalized radial frequency, with the DC neighborhood clamped to a small positive value for numerical stability. Equation~\eqref{eq:ideal_powers} transfers the ideal rectified-flow marginal reliability structure to the endpoint discrepancy, and it is a proxy rather than an identity for the true discrepancy powers. Since $S_\beta$ depends only on radial frequency, it is a radially isotropic power-law approximation to the latent's generally anisotropic spectrum, capturing the dominant decay without image-specific spectral estimation. We use $\beta=2$, consistent with the approximate power-law behavior of natural-image spectra~\citep{field1987relations}, and verify the corresponding latent-space trend in \suppref{app:spectrum}.
Substituting Eq.~\eqref{eq:ideal_powers} into Eq.~\eqref{eq:oracle_wiener} gives
\begin{equation}
    W_{\mathrm{ideal}}(t,f)
    =
    \frac{
        (1-t)^2S_\beta(f)
    }{
        (1-t)^2S_\beta(f)+t^2
    }.
    \label{eq:ideal_wiener}
\end{equation}

\paragraph{Learned-flow uncertainty generalization.}
The $t^2$ term in Eq.~\eqref{eq:ideal_powers} assumes a spectrally white velocity-prediction error of constant power. However, the endpoint accuracy of a pretrained flow varies across timesteps and frequencies, and target-side CFG adds further endpoint mismatch. We therefore preserve the Wiener ratio but replace the ideal white uncertainty with
\begin{equation}
    N_\phi(t,f) = n_0\,t^{\tau}r(f)^{-\nu},
    \qquad
    \phi=(n_0,\tau,\nu).
    \label{eq:noise_model}
\end{equation}
Here $n_0$ sets the overall error power, $\tau$ controls how quickly the endpoint uncertainty grows with the noise level $t$, and $\nu$ tilts the error spectrum toward the low frequencies where prediction bias and CFG-guided extrapolation are expected to concentrate. Frequency then enters the gain only through $\nu-\beta$, so the correction favors low frequencies whenever the source prior decays faster than the error coloring and becomes frequency-flat when the two coincide, the regime scalar correction schedules assume  (see \suppref{app:proxy_derivation}). We fix $\phi$ globally across images and editing instructions and analyze the corresponding empirical spectra in \suppref{app:spectrum}. The generalized frequency-wise gain is therefore
\begin{equation}
    W_{\beta,\phi}(t,f)
    =
    \frac{(1-t)^2S_\beta(f)}{(1-t)^2S_\beta(f)+N_\phi(t,f)},
    \label{eq:wiener_gain}
\end{equation}
the exact componentwise linear-MMSE gain under the adopted frequency-diagonal, channel-isotropic proxy power model. Setting $\beta=\nu=0$ and $\phi=(1,2,0)$ recovers the ideal rectified-flow SNR schedule $W(t)=(1-t)^2/\bigl((1-t)^2+t^2\bigr)$.

\paragraph{Spectral normalization and correction.}
Different $(\beta,\phi)$ change both the spectral shape and the RMS
magnitude of the gain map. To compare spectral variants at a matched
gain-map scale, we divide the gain by its RMS value over the active
timestep set $\mathcal I$ and frequency grid $\Omega$:
\begin{equation}
    C_W
    =
    \left[
        \frac{1}{|\mathcal I||\Omega|}
        \sum_{t_i\in\mathcal I}
        \sum_{f\in\Omega}
        W_{\beta,\phi}(t_i,f)^2
    \right]^{1/2}.
    \label{eq:budget_norm}
\end{equation}
The filtered discrepancy and source-consistency correction proposal are
\begin{equation}
\begin{aligned}
    \bar\delta_i^{\,w}
    &=
    \mathcal F^{-1}
    \left\{
        \frac{
            W_{\beta,\phi}(t_i,f)
        }{
            C_W
        }
        \mathcal F\{\delta_i\}
    \right\},\\
    u_i^{\mathrm{corr}}
    &=
    \alpha\bar\delta_i^{\,w},
\end{aligned}
\label{eq:correction_proposal}
\end{equation}
where $\mathcal F$ is a per-channel two-dimensional Fourier transform,
the scalar gain is shared across latent channels, and $\alpha$ controls
the global correction strength. The normalization fixes the RMS scale
of the gain map itself, rather than assuming a white spectrum for every
input discrepancy (see \suppref{app:normalization_derivation}).

\subsection{Energy-Guided Soft Gate Decomposition (SGD)}
\label{sec:decomposition}
The frequency-wise source-consistency correction in Eq.~\eqref{eq:correction_proposal} is spatially shift-invariant, whereas text-guided editing requires different behavior across spatial locations. We therefore interpret the target-directed editing and source-consistency correction streams as competing local update proposals and allocate their responsibility using model-derived energies. For a $d$-channel latent tensor $x$, we define the channel-wise RMS at
each spatial location $p$ as
\begin{equation}
    \operatorname{RMS}_c(x)(p)
    =
    \left[
        \frac{1}{d}
        \sum_{k=1}^{d}
        x_k(p)^2
        +
        \varepsilon
    \right]^{1/2},
    \label{eq:channel_rms}
\end{equation}
where $\varepsilon$ is a small positive constant that stabilizes this and all subsequent divisions.

\paragraph{Stabilized source-orthogonal editing evidence.}
The magnitude of $u_i^{\mathrm{edit}}$ alone is not necessarily reliable evidence for local semantic editing, since part of the target velocity may follow the source branch's own direction. Therefore at each spatial location $p$ we compute
\begin{equation}
\begin{aligned}
    v_i^{\mathrm{tar},\perp}(p)
    &=
    \widetilde v_i^{\mathrm{tar}}(p)
    -
    \frac{
        \left\langle
            \widetilde v_i^{\mathrm{tar}}(p),
            \widetilde v_i^{\mathrm{src}}(p)
        \right\rangle_c
    }{
        \max\left(
            \left\|
                \widetilde v_i^{\mathrm{src}}(p)
            \right\|_c^2,
            \varepsilon
        \right)
    }
    \widetilde v_i^{\mathrm{src}}(p),\\
    E_i^{\mathrm{edit}}(p)
    &=
    \operatorname{RMS}_c
    \left(
        \Delta t_i
        v_i^{\mathrm{tar},\perp}
    \right)(p).
\end{aligned}
\label{eq:edit_energy}
\end{equation}
The inner product and norm are taken over latent channels at each spatial location. When the source-velocity norm exceeds the stabilization threshold, Eq.~\eqref{eq:edit_energy} measures local motion orthogonal to the source flow, and the clamped denominator provides a stable approximation in near-zero-norm regions. This energy serves only as gating evidence, while the actual editing proposal remains Eq.~\eqref{eq:edit_proposal}.

\paragraph{Source-consistency correction evidence.}
For the source-consistency correction stream, we measure the
normalized Wiener-filtered discrepancy before applying $\alpha$:
\begin{equation}
    E_i^{\mathrm{corr}}(p)
    =
    \operatorname{RMS}_c
    \left(
        \bar\delta_i^{\,w}
    \right)(p).
    \label{eq:corr_energy}
\end{equation}
Because $u_i^{\mathrm{corr}}=\alpha\bar\delta_i^{\,w}$, we use
$\alpha E_i^{\mathrm{corr}}(p)$ as a numerically stabilized proxy for
the local magnitude of the applied correction proposal.
The two local energies define the editing-responsibility field
\begin{equation}
    G_i(p)
    =
    \frac{
        E_i^{\mathrm{edit}}(p)
    }{
        E_i^{\mathrm{edit}}(p)
        +
        \alpha E_i^{\mathrm{corr}}(p)
        +
        \varepsilon
    }.
    \label{eq:soft_gate}
\end{equation}
A large $G_i(p)$ favors target-directed editing, whereas a small value
favors source-consistency correction.
The edited latent is updated by
\begin{equation}
    Z_{i-1}^{\mathrm{edit}}
    =
    Z_i^{\mathrm{edit}}
    +
    G_i\odot u_i^{\mathrm{edit}}
    +
    (1-G_i)\odot u_i^{\mathrm{corr}},
    \label{eq:full_update}
\end{equation}
where $\odot$ denotes elementwise multiplication and the single-channel gate is broadcast over latent channels. Rather than superimposing the correction on the editing update, Eq.~\eqref{eq:full_update} lets the two proposals compete locally on model-derived evidence. It requires no segmentation mask, auxiliary network, or per-image optimization, and adds no tunable coefficient beyond the $\Delta t_i$ and $\alpha$ that already set the proposal magnitudes.

Algorithm~\ref{alg:method} summarizes the main FCEdit pipeline. In Algorithm~\ref{alg:method}, $\mathcal V_\theta(z_i^{\mathrm{src}},z_i^{\mathrm{tar}})$ denotes the single batched CFG evaluation that returns $(\widetilde v_i^{\mathrm{src}}, \widetilde v_i^{\mathrm{tar}}, v_i^{\mathrm{src},c})$. The explicit four-branch batched evaluation and implementation-level details are provided in \suppref{app:full_algorithm}.

\begin{figure*}[t]
    \centering
    \setlength{\tabcolsep}{1pt}
    \renewcommand{\arraystretch}{0.75}
    \resizebox{0.98\textwidth}{!}{%
    \begin{tikzpicture}
    \node[inner sep=0pt] (qualgrid) {\begin{tabular}{ccccccccccc}
        \multicolumn{6}{c}{\textbf{SD3.5}} & & \multicolumn{4}{c}{\textbf{FLUX}} \\
        Source & \textbf{FCEdit} & DRFS~\citep{drfs} & FIA-Edit~\citep{fia_edit} & FlowAlign~\citep{flow_align} & FlowEdit~\citep{flow_edit} & &
        Source & \textbf{FCEdit} & FlowEdit~\citep{flow_edit} & RF-Inv~\citep{Rout2024SemanticII} \\
        \qualimg{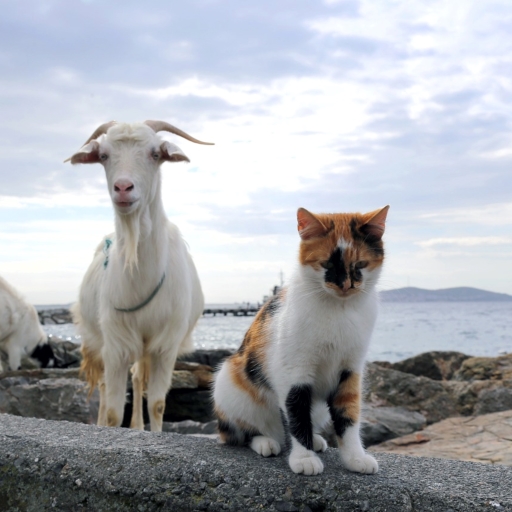} &
        \qualimg{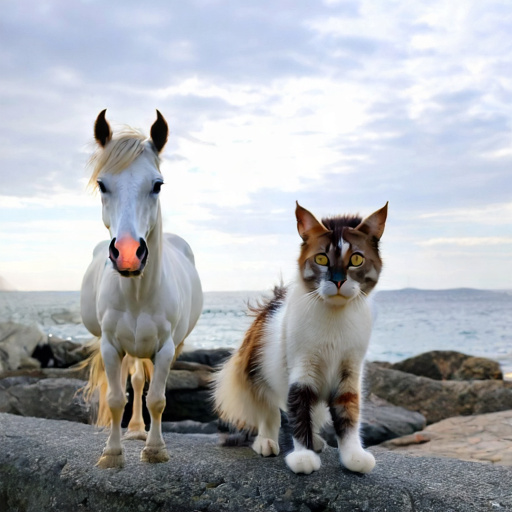} &
        \qualcirc{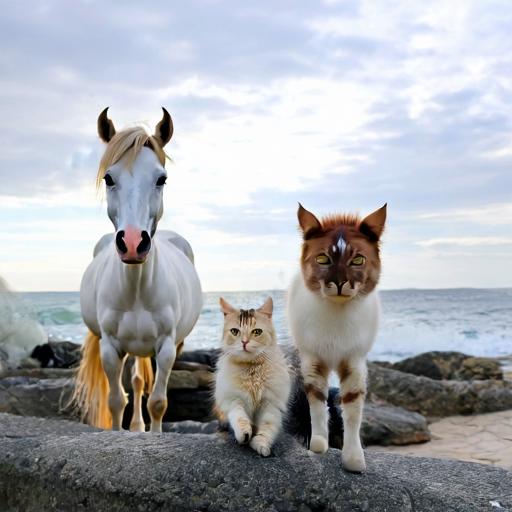}{0.48}{0.24}{0.2}{0.2} &
        \qualcirc{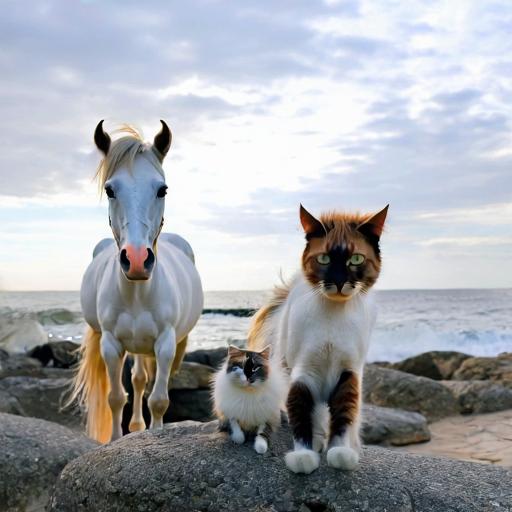}{0.48}{0.24}{0.2}{0.2} &
        \qualcirc{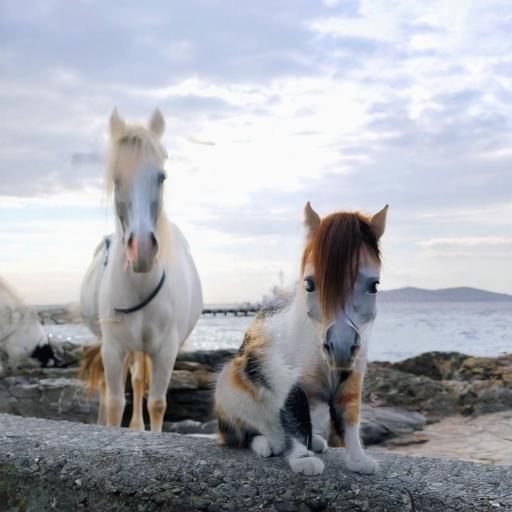}{0.62}{0.36}{0.2}{0.32} &
        \qualcirc{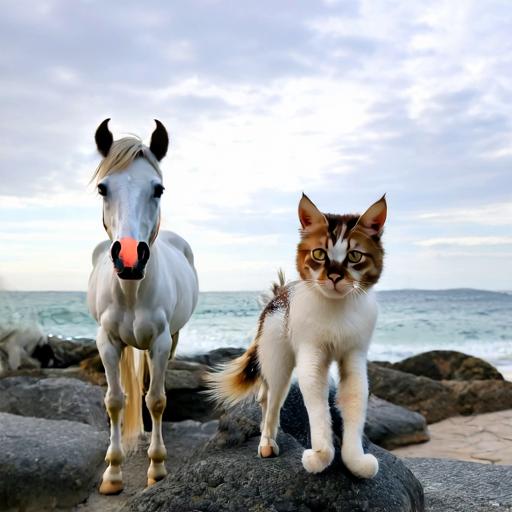}{0.52}{0.24}{0.19}{0.2} & &
        \qualimg{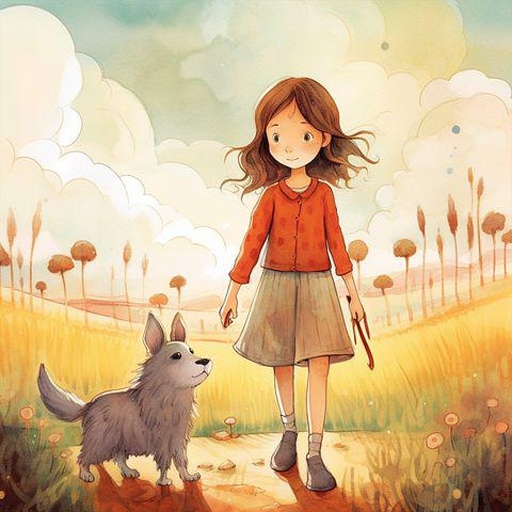} &
        \qualimg{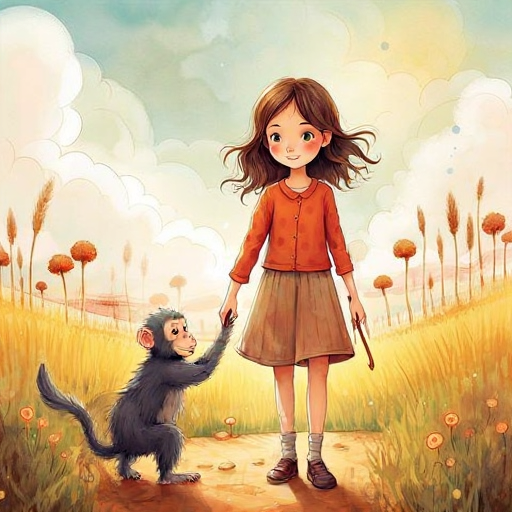} &
        \qualcirc{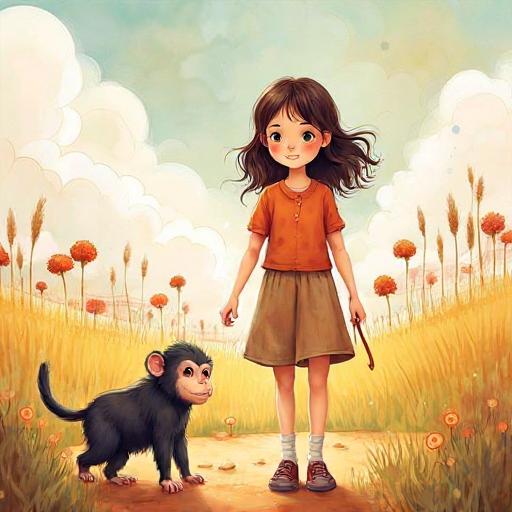}{0.62}{0.41}{0.19}{0.2} &
        \qualcirc{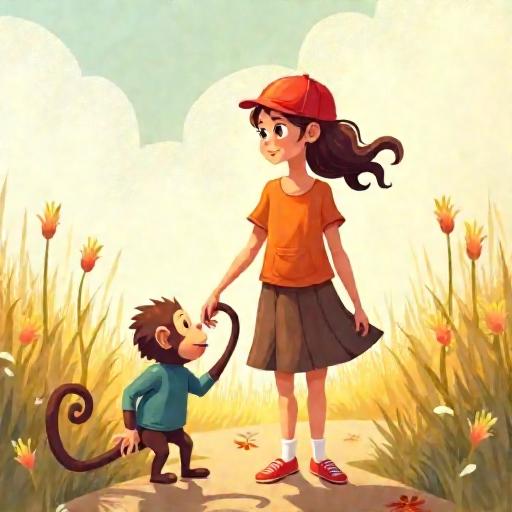}{0.64}{0.68}{0.28}{0.29} \\
        \multicolumn{6}{c}{\scalebox{0.8}{\large\sffamily\textit{goat} $\rightarrow$ \textcolor{qualpurple}{\large\sffamily\textit{horse}}}} & &
        \multicolumn{4}{c}{\scalebox{0.8}{\large\sffamily\textit{dog} $\rightarrow$ \textcolor{qualpurple}{\large\sffamily\textit{monkey}}}} \\[0pt]
        \qualimg{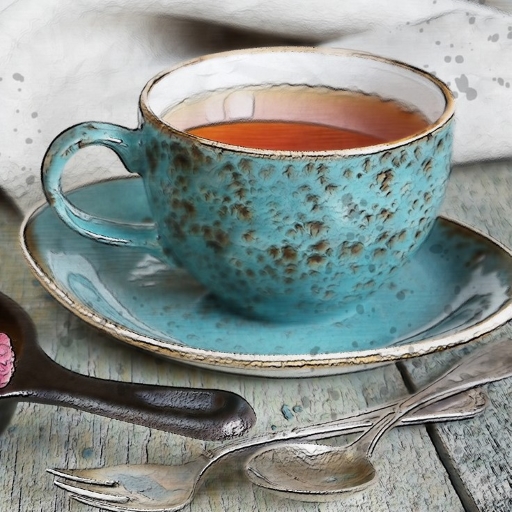} &
        \qualimg{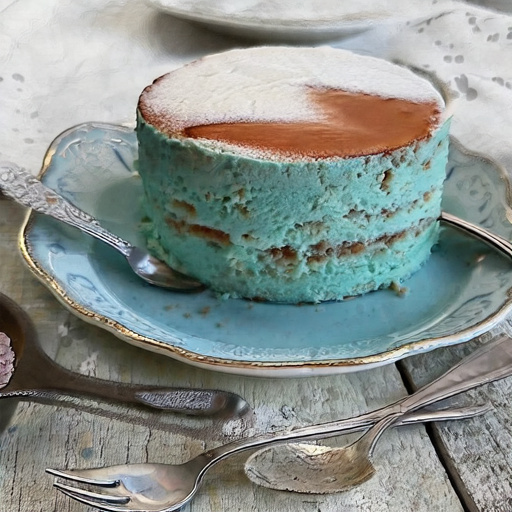} &
        \qualcirc{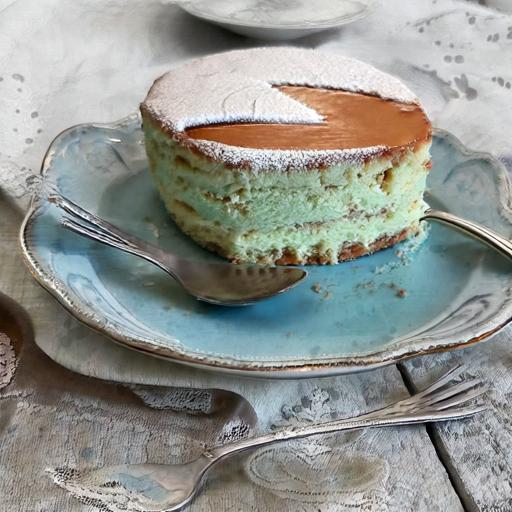}{0.25}{0.25}{0.2}{0.21} &
        \qualcirc{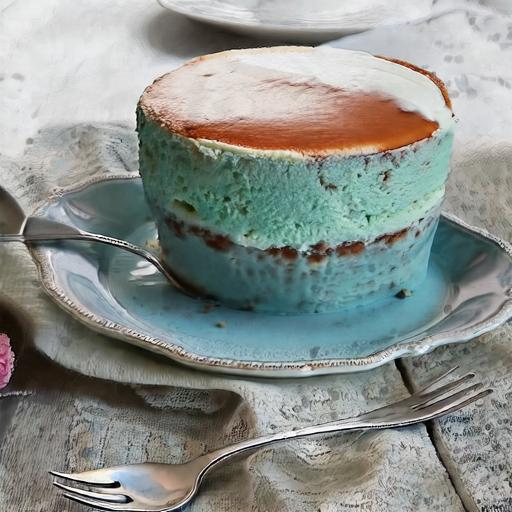}{0.76}{0.25}{0.2}{0.21} &
        \qualimg{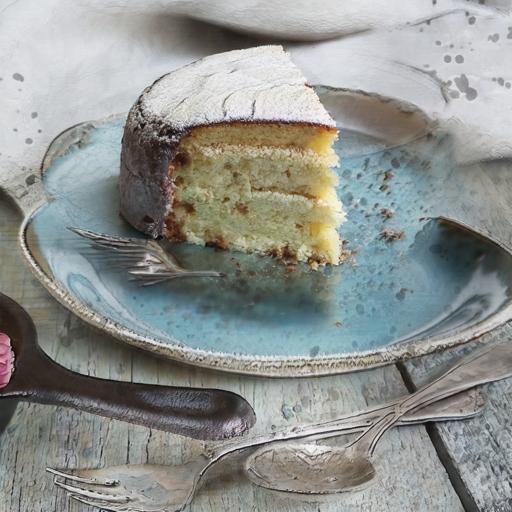} &
        \qualcirc{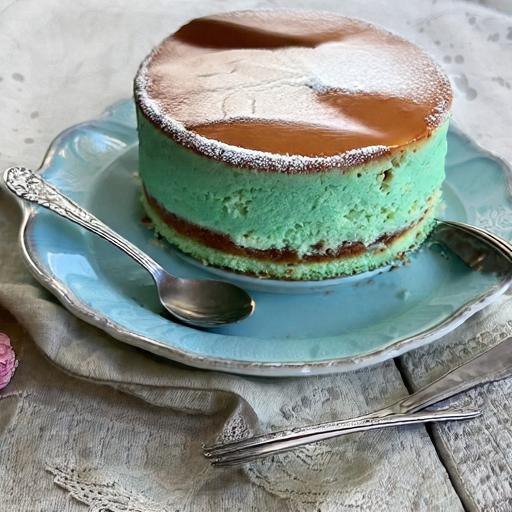}{0.31}{0.25}{0.2}{0.21} & &
        \qualimg{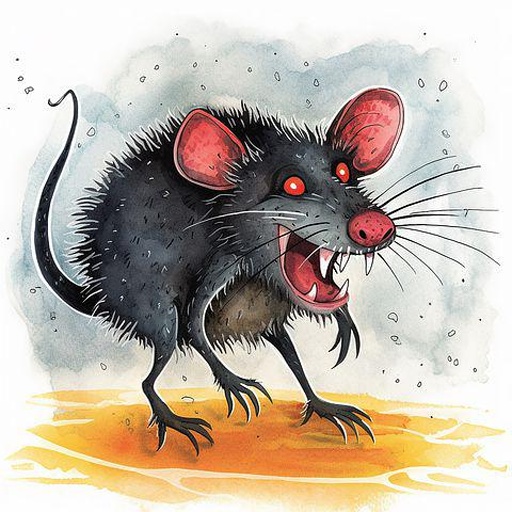} &
        \qualimg{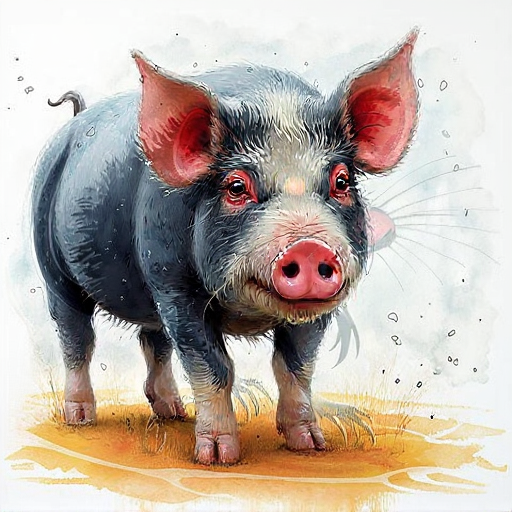} &
        \qualimg{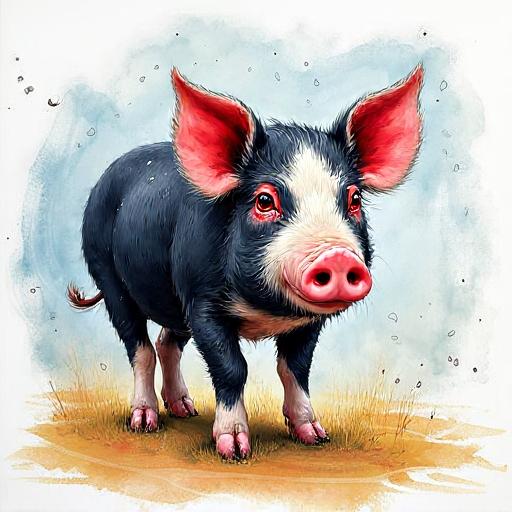} &
        \qualimg{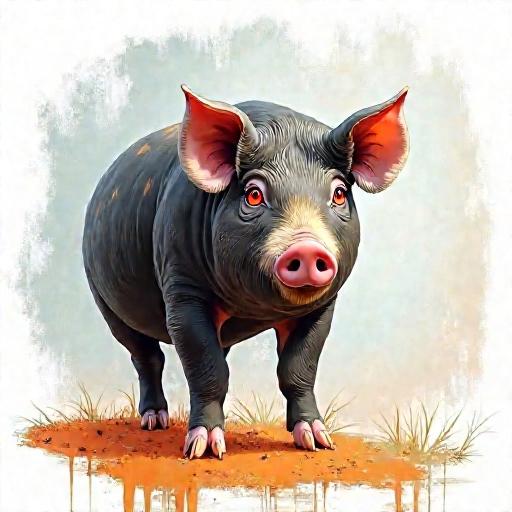} \\
        \multicolumn{6}{c}{\scalebox{0.8}{\large\sffamily\textit{teacup} $\rightarrow$ \textcolor{qualpurple}{\large\sffamily\textit{cake}}}} & &
        \multicolumn{4}{c}{\scalebox{0.8}{\large\sffamily\textit{rat} $\rightarrow$ \textcolor{qualpurple}{\large\sffamily\textit{pig}}}} \\[0pt]
        \qualimg{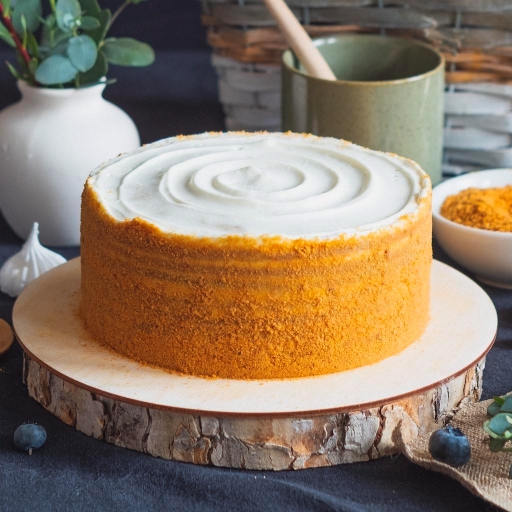} &
        \qualimg{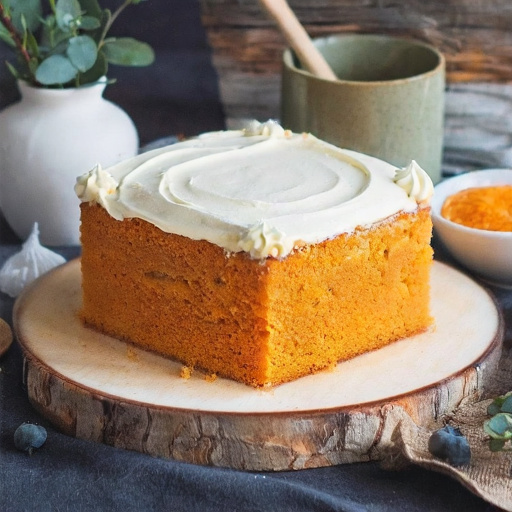} &
        \qualcirc{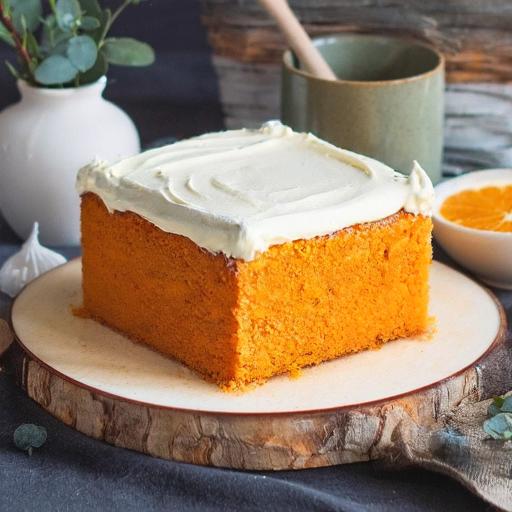}{0.82}{0.18}{0.16}{0.15} &
        \qualimg{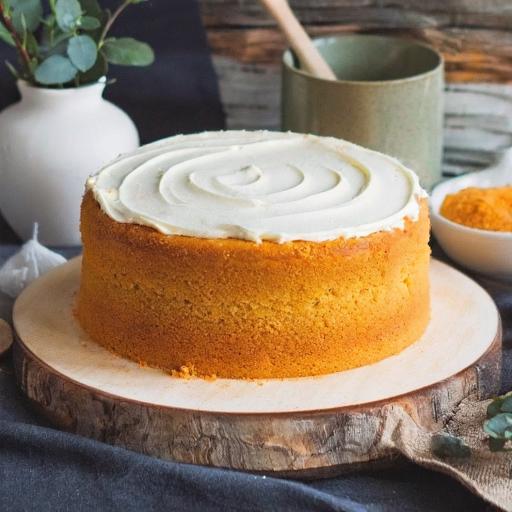} &
        \qualcirc{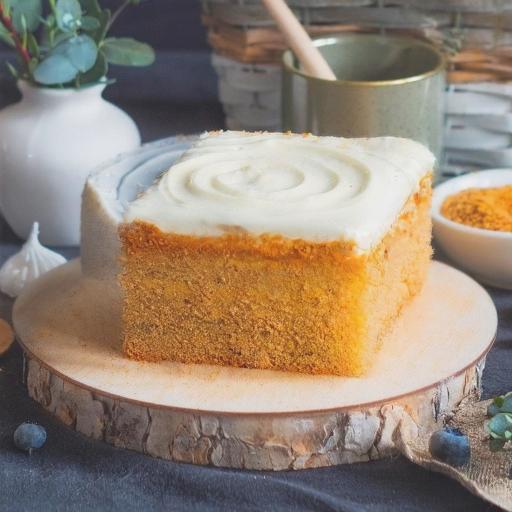}{0.27}{0.6}{0.18}{0.2} &
        \qualcirc{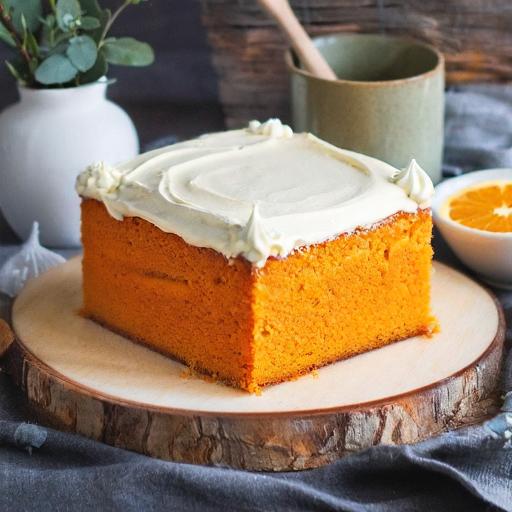}{0.78}{0.18}{0.17}{0.15} & &
        \qualimg{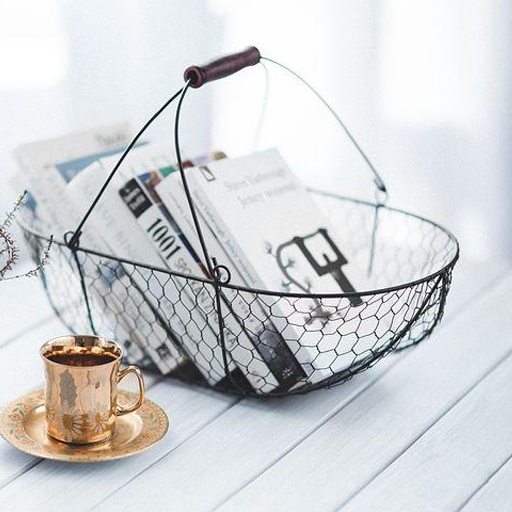} &
        \qualimg{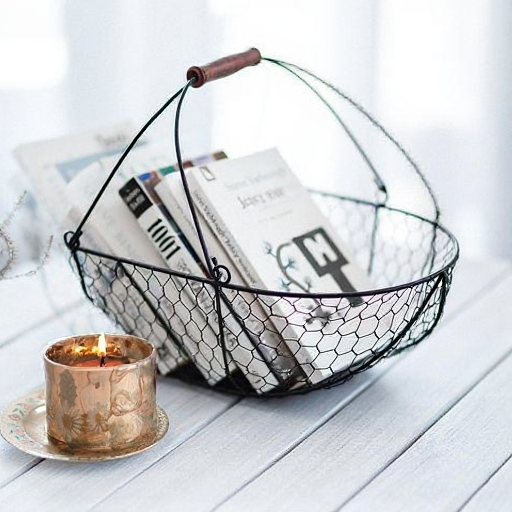} &
        \qualcirc{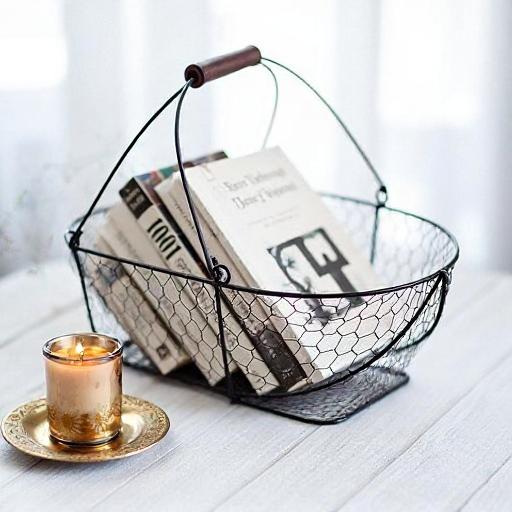}{0.42}{0.49}{0.2}{0.19} &
        \qualimg{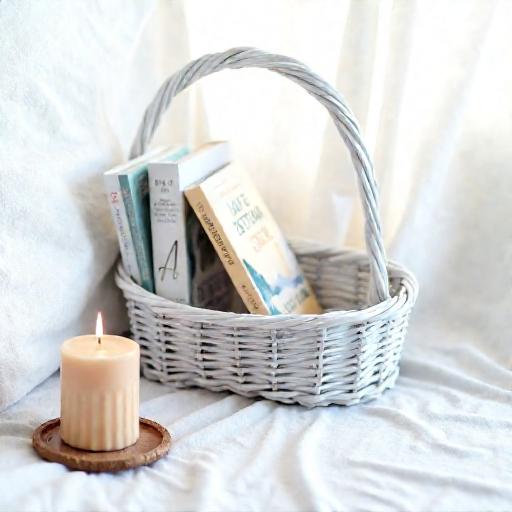} \\
        \multicolumn{6}{c}{\scalebox{0.8}{\large\sffamily\textit{round} $\rightarrow$ \textcolor{qualpurple}{\large\sffamily\textit{square}}}} & &
        \multicolumn{4}{c}{\scalebox{0.8}{\large\sffamily\textit{cup} $\rightarrow$ \textcolor{qualpurple}{\large\sffamily\textit{candle}}}} \\[0pt]
        \qualimg{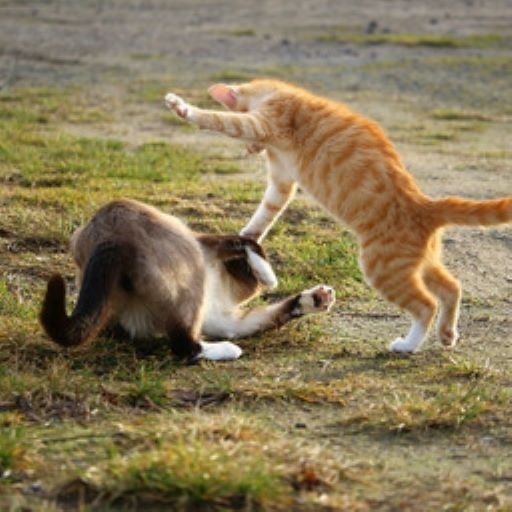} &
        \qualimg{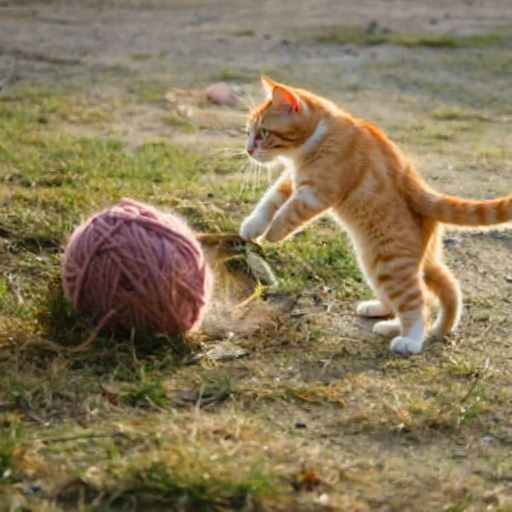} &
        \qualcirc{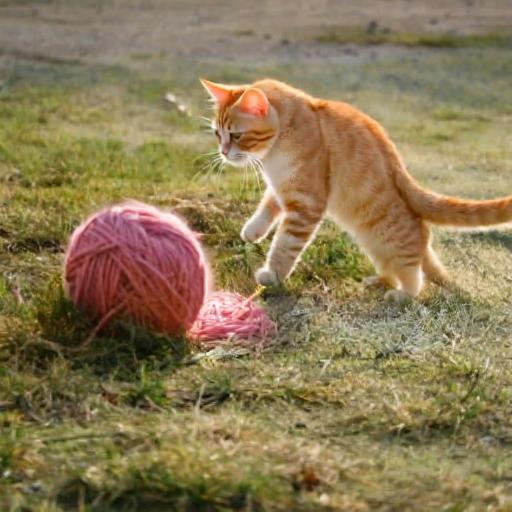}{0.5}{0.34}{0.17}{0.17} &
        \qualcirc{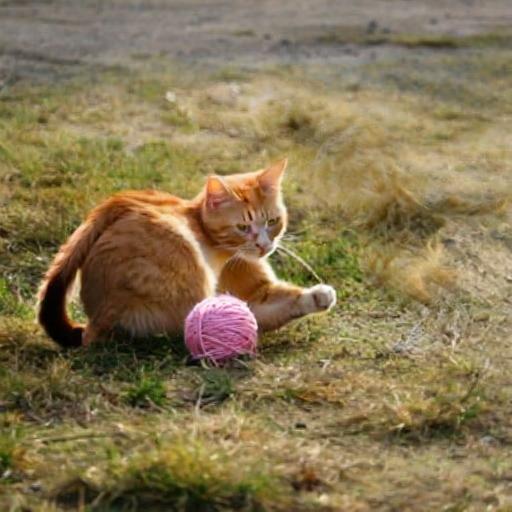}{0.75}{0.7}{0.2}{0.25} &
        \qualimg{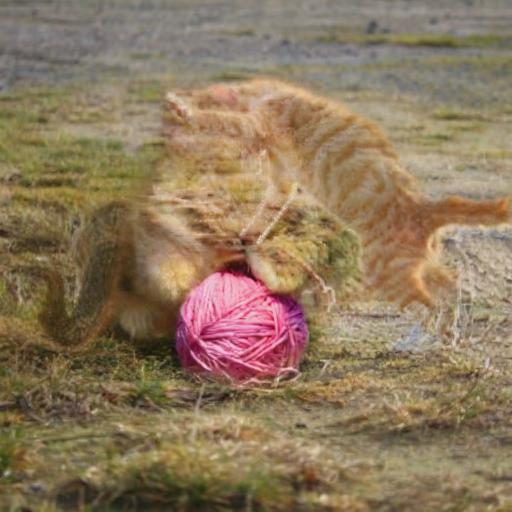} &
        \qualimg{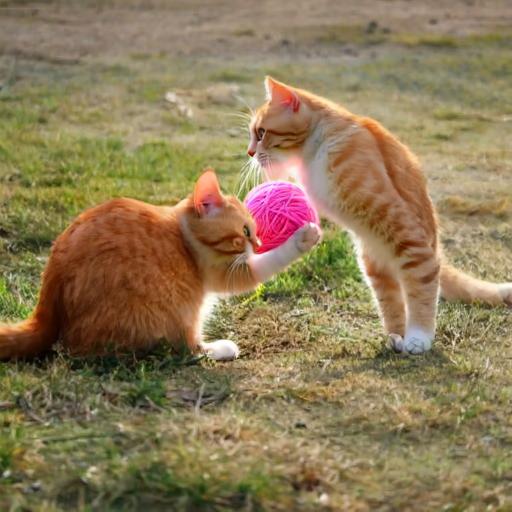} & &
        \qualimg{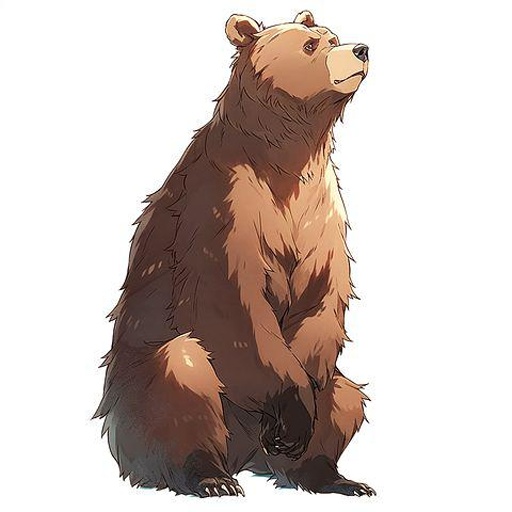} &
        \qualimg{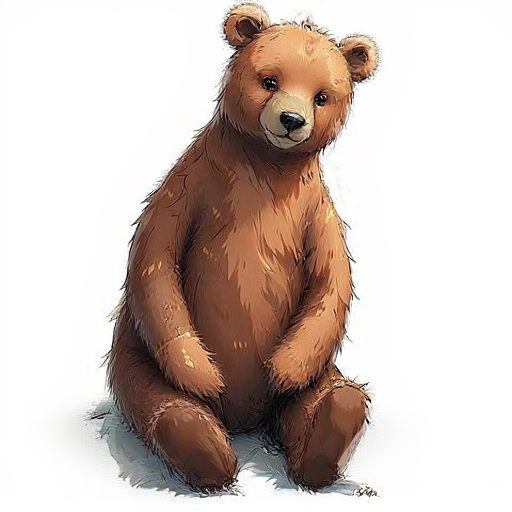} &
        \qualimg{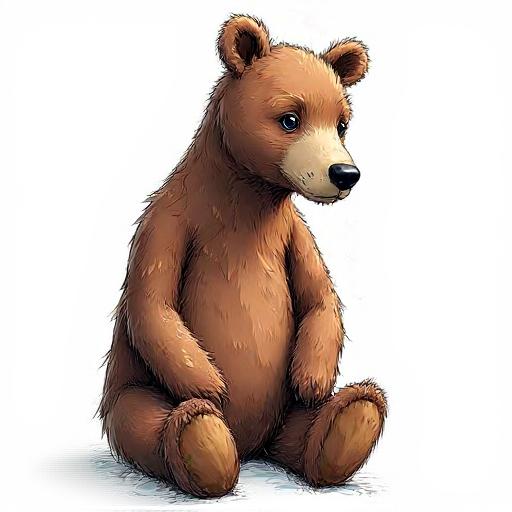} &
        \qualimg{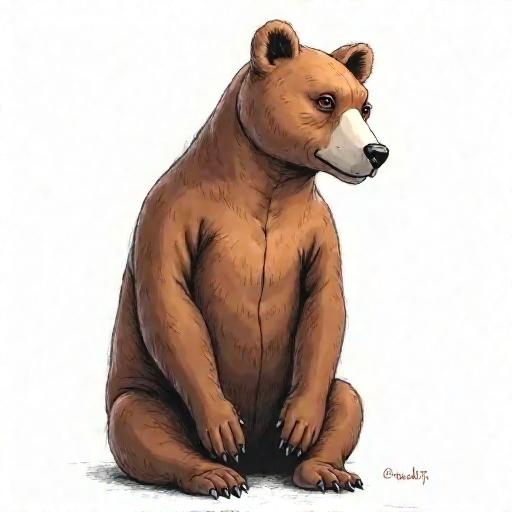} \\
        \multicolumn{6}{c}{\scalebox{0.8}{\large\sffamily\textit{black and white cat} $\rightarrow$ \textcolor{qualpurple}{\large\sffamily\textit{pink yarn ball}}}} & &
        \multicolumn{4}{c}{\scalebox{0.8}{\large\sffamily\textit{brown bear} $\rightarrow$ \textcolor{qualpurple}{\large\sffamily\textit{knitted brown bear toy}}}} \\
    \end{tabular}};
    \draw[black,dash pattern=on 1pt off 1pt,line width=0.5pt] ($(qualgrid.south west)!0.1!(qualgrid.south east)+(0,14pt)$) -- ($(qualgrid.north west)!0.1!(qualgrid.north east)-(0,24pt)$);
    \draw[black,solid,line width=0.5pt] ($(qualgrid.south west)!0.601!(qualgrid.south east)+(0,14pt)$) -- ($(qualgrid.north west)!0.601!(qualgrid.north east)-(0,24pt)$);
    \draw[black,dash pattern=on 1pt off 1pt,line width=0.5pt] ($(qualgrid.south west)!0.701!(qualgrid.south east)+(0,14pt)$) -- ($(qualgrid.north west)!0.701!(qualgrid.north east)-(0,24pt)$);
    \end{tikzpicture}%
    }
    \caption{
        \textbf{Qualitative comparisons on SD3.5 (left) and FLUX (right).} We compare FCEdit with the inversion-free methods FlowEdit~\citep{flow_edit}, FlowAlign~\citep{flow_align}, FIA-Edit~\citep{fia_edit}, and DRFS~\citep{drfs}, and with the inversion-based method RF-Inversion~\citep{Rout2024SemanticII}. FCEdit achieves semantically aligned edits while better preserving the source structure in non-edited regions. The red dashed circles indicate incomplete edits, structural artifacts, and unintended modifications.}
    \label{fig:qual}
\end{figure*}
\input{table/quant}

\input{table/algo}

\paragraph{Default configuration.}
Unless otherwise stated, we use a full discretization of $T=50$
rectified-flow steps and set the active FCEdit subsequence to the final
$N=35$ low-noise steps. We set
$\eta_{\mathrm{src}}=3.5$, $\eta_{\mathrm{tar}}=15.5$,
$\beta=2$, $r_{\mathrm{ref}}=1$,
$\phi=(n_0,\tau,\nu)=(1,1,0.2)$, and
$\alpha=0.008$. The editing evidence is computed from the
source-orthogonal target velocity, while the correction evidence is
measured before applying $\alpha$. All parameters are fixed across
all experiments.

%% file: table/quant.tex
\begin{table*}[t]
    \centering
    \setlength{\tabcolsep}{6.5pt}
    \renewcommand{\arraystretch}{0.90}
    \resizebox{\textwidth}{!}{%
    \begin{tabular}{@{}llcccccccc@{}}
        \toprule
        & & 
        & \multicolumn{1}{c}{Structure}
        & \multicolumn{4}{c}{Background Preservation}
        & \multicolumn{2}{c}{CLIP Similarity} \\
        \cmidrule(lr){4-4} \cmidrule(lr){5-8} \cmidrule(lr){9-10}
        Method & Model & Inv-F
        & Distance $\times 10^{3}\!\downarrow$
        & PSNR $\uparrow$
        & LPIPS $\times 10^{3}\!\downarrow$
        & MSE $\times 10^{4}\!\downarrow$
        & SSIM $\times 10^{2}\!\uparrow$
        & Whole $\uparrow$
        & Edited $\uparrow$ \\
        \midrule
        \textit{FCEdit (Ours)} & SD3.5 & \cmark
        & \textbf{10.14} & \textbf{27.35} & \textbf{48.40} & \underline{29.49}
& \textbf{90.03} & \textbf{26.95} & 23.26$^{\dagger}$ \\
        \midrule
        FlowEdit~\cite{flow_edit} & SD3.5 & \cmark
        & 23.45 & 23.24 & 93.51 & 69.90
        & 85.12 & \underline{26.80} & \textbf{23.69} \\
        FlowAlign~\cite{flow_align} & SD3.5 & \cmark
        & 31.15 & 24.60 & 64.61$^{\dagger}$ & 48.99
        & 86.76 & 25.70 & 22.40 \\
        FIA-Edit~\cite{fia_edit} & SD3.5 & \cmark
        & \underline{10.34} & \underline{27.31} & \underline{55.07} & \textbf{28.66}
        & \underline{89.20} & 25.90 & 22.78 \\
        DRFS~\cite{drfs} & SD3.5 & \cmark
        & 14.36$^{\dagger}$ & 25.99$^{\dagger}$ & 64.83 & 38.04$^{\dagger}$
        & 88.36$^{\dagger}$ & 26.46$^{\dagger}$ & \underline{23.36} \\
        \midrule
        \textit{FCEdit (Ours)} & FLUX & \cmark
        & \textbf{14.42} & \textbf{25.19} & \textbf{64.94} & \textbf{46.83}
& \textbf{89.27} & \underline{25.42} & 22.01$^{\dagger}$ \\
        \midrule
        RF-Inversion~\cite{Rout2024SemanticII} & FLUX & \xmark
        & 64.45 & 17.65 & 249.99 & 231.18
& 64.68 & 25.23 & 22.84 \\
        FlowEdit~\cite{flow_edit} & FLUX & \cmark
        & 28.36 & 21.95 & 108.45 & 95.39
& 84.14 & \textbf{26.18} & 22.89 \\
        
        \bottomrule
    \end{tabular}%
    }
    \caption{\textbf{Quantitative comparison} on PIE-Bench with 700 images.
    Inv-F denotes methods that do not estimate a source-specific noise or
    latent representation in a separate pre-editing stage. PSNR, LPIPS, MSE,
    and SSIM are evaluated over the background region of the 556 pairs with a non-empty background mask. Lower is better for
    Distance, LPIPS, and MSE, and higher is better for the remaining metrics.
    Rankings are computed over the SD3.5 rows, and the FLUX rows are reported
    as a backbone reference. The best, second-best, and third-best scores are
    shown in \textbf{bold}, \underline{underline}, and $\dagger$, respectively.}
    \label{tab:main_results}
\end{table*}

%% file: table/algo.tex
\begin{algorithm}[tb]
\caption{FCEdit Pipeline}
\label{alg:method}
\small
\begin{algorithmic}[1]
\REQUIRE
source image $I_{\mathrm{src}}$, prompts $c_{\mathrm{src}},c_{\mathrm{tar}}$,
flow model $v_\theta$, autoencoder $(\mathcal E,\mathcal D)$,
active timesteps $\{t_i\}_{i=0}^{N}$,
guidance scales $\eta_{\mathrm{src}},\eta_{\mathrm{tar}}$,
gain parameters $\Theta_W$, correction strength $\alpha$
\ENSURE edited image $I_{\mathrm{edit}}$
\STATE $x_{\mathrm{src}}\leftarrow\mathcal E(I_{\mathrm{src}})$;
\quad $Z_N^{\mathrm{edit}}\leftarrow x_{\mathrm{src}}$
\STATE precompute the normalized gain $W_i(f)/C_W$
\hfill $\triangleright$ Eqs.~\eqref{eq:wiener_gain}, \eqref{eq:budget_norm}
\FOR{$i=N,\ldots,1$}
    \STATE sample $\epsilon_i\sim\mathcal N(0,I)$ and
    build $z_i^{\mathrm{src}},z_i^{\mathrm{tar}}$
    \hfill $\triangleright$ Eq.~\eqref{eq:flowedit_states}
    \STATE $(\widetilde v_i^{\mathrm{src}},
      \widetilde v_i^{\mathrm{tar}},
      v_i^{\mathrm{src},c})
    \leftarrow
    \mathcal V_\theta(z_i^{\mathrm{src}},z_i^{\mathrm{tar}})$
    \hfill $\triangleright$ one batched pass
    \STATE $u_i^{\mathrm{edit}}
    \leftarrow
    \Delta t_i(\widetilde v_i^{\mathrm{tar}}-\widetilde v_i^{\mathrm{src}})$
    \hfill $\triangleright$ editing proposal, Eq.~\eqref{eq:edit_proposal}
    \STATE $\delta_i
    \leftarrow
    x_{\mathrm{src}}-Z_i^{\mathrm{edit}}
    +t_i(\widetilde v_i^{\mathrm{tar}}-v_i^{\mathrm{src},c})$
    \hfill $\triangleright$ discrepancy, Eq.~\eqref{eq:discrepancy_decomposition}
    \STATE $\bar\delta_i^{\,w}
    \leftarrow
    \mathcal F^{-1}\{(W_i/C_W)\,\mathcal F\delta_i\},
    \;
    u_i^{\mathrm{corr}}\leftarrow\alpha\bar\delta_i^{\,w}$
    \hfill $\triangleright$ Eq.~\eqref{eq:correction_proposal}
    \STATE compute evidence energies
    $E_i^{\mathrm{edit}},E_i^{\mathrm{corr}}$
    \hfill $\triangleright$ Eqs.~\eqref{eq:edit_energy}, \eqref{eq:corr_energy}
    \STATE $G_i
    \leftarrow
    E_i^{\mathrm{edit}}/
    (E_i^{\mathrm{edit}}+\alpha E_i^{\mathrm{corr}}+\varepsilon)$
    \hfill $\triangleright$ soft gate, Eq.~\eqref{eq:soft_gate}
    \STATE $Z_{i-1}^{\mathrm{edit}}
    \leftarrow
    Z_i^{\mathrm{edit}}
    +G_i\odot u_i^{\mathrm{edit}}
    +(1-G_i)\odot u_i^{\mathrm{corr}}$
    \hfill $\triangleright$ Eq.~\eqref{eq:full_update}
\ENDFOR
\RETURN $I_{\mathrm{edit}}=\mathcal D(Z_0^{\mathrm{edit}})$
\end{algorithmic}
\end{algorithm}

%% file: sec/4_exp.tex
\section{Experiments}
\label{sec:exp}

\subsection{Experimental Setup}
We evaluate text-guided image editing on PIE-Bench~\citep{pie_bench}, which comprises 700 images spanning nine editing categories. We compare FCEdit against the inversion-based RF-Inversion~\citep{Rout2024SemanticII} and inversion-free FlowEdit~\citep{flow_edit}, FlowAlign~\citep{flow_align}, FIA-Edit~\citep{fia_edit}, and DRFS~\citep{drfs}, each run with its officially recommended hyperparameters. We report Structure Distance~\cite{tumanyan2022splicing,caron2021dino} for structural consistency, LPIPS~\cite{Zhang2018TheUE}, SSIM~\cite{Wang2004ImageQA}, MSE, and PSNR outside the annotated edit mask for background preservation, and CLIP similarity~\cite{Radford2021LearningTV} on the whole image (CLIP-Whole) and on the edited region (CLIP-Edited) for text--image alignment. We further conduct a human preference study against the inversion-free methods. Implementation details are provided in \suppref{app:implementation}. We also provide additional comparisons with recent flow editors~\cite{xiednaedit,avrahami2025stableflow,dengfireflow,wang2025taming} and few- or one-step editors~\cite{wu2024turboedit,chord_edit,  nguyen2025swiftedit} 
in \suppref{app:extra_baselines}.

\subsection{Experimental Results}

\paragraph{Quantitative comparisons.}
Table~\ref{tab:main_results} reports quantitative results on PIE-Bench. 
FCEdit achieves the best structure distance, PSNR, LPIPS, and SSIM. It also achieves the highest whole-image CLIP similarity while remaining competitive on edited-region CLIP similarity. FCEdit outperforms FIA-Edit, the strongest source-preservation baseline, reducing background LPIPS by 12\% and also surpasses FlowEdit, our strong inversion-free baseline, in CLIP-Whole. Against FlowEdit, which follows the same direct-editing trajectory without correction, FCEdit more than halves the structure distance (10.14 vs.\ 23.45) while giving up only 0.43 points of edited-region CLIP score (23.26 vs.\ 23.69). Against FlowAlign, which feeds back the same endpoint correction at a constant strength, FCEdit improves source-preservation metrics and achieves higher CLIP scores. Overall, FCEdit achieves markedly stronger source preservation at competitive editability. 

\input{table/quant_user}

\paragraph{Qualitative comparisons.}
As shown in Figure~\ref{fig:qual}, FCEdit produces semantically correct edits within the intended regions while naturally preserving the source background, details, and style. The red dashed circles indicate misplaced edits, unintended generation, or failures to preserve non-edited content. In the first and fourth of SD3.5 examples, competing methods frequently place the newly generated object at an incorrect location, overwrite a nearby object, or alter a larger region than required by the instruction. In contrast, FCEdit confines the modification to the intended area and better maintains the surrounding scene composition. In the second row, FCEdit is the only method that fully preserves the objects surrounding the teacup, whereas other methods remove or deform nearby content despite its irrelevance to the target instruction. In the cake example, FIA-Edit fails to complete the requested shape change, while several competing methods also remove the blueberries that should remain unchanged. These cases illustrate that strong semantic generation alone does not guarantee a precise edit, particularly when the target region is spatially close to source content that must be preserved.
On FLUX, FCEdit similarly preserves source details across changes in object identity, material, and texture. In the first row, FCEdit better preserves the girl’s appearance and pose while changing the dog into a monkey. In the third row, it retains the surrounding scene composition while replacing the cup with a candle.
These results suggest that the source-preservation benefit of FCEdit extends across both flow backbones.

\paragraph{Human preference study.}
We also conducted a human preference study with 25 completion-filtered Prolific participants~\cite{prolific2026}. Each participant evaluated 24 instances drawn from 50 PIE-Bench examples, selecting up to two preferred results among five anonymized methods for each criterion. As shown in Table~\ref{tab:human_pref}, FCEdit ranks first on semantic alignment, source consistency, and overall preference, with selection rates of 37.8\%, 36.8\%, and 38.3\% respectively.

\begin{figure}[!t]
    \centering
    \setlength{\tabcolsep}{1pt}
    \renewcommand{\arraystretch}{0.95}
    \def\ablationprompt#1{{\large\sffamily #1}}
    \def\ablationgrid{%
    \begin{tabular}{cccc}
        \makebox[\qualw][c]{\fontsize{7}{8}\selectfont Source} &
        \makebox[\qualw][c]{\fontsize{7}{8}\selectfont Edit Only} &
        \makebox[\qualw][c]{\fontsize{7}{8}\selectfont +Correction ($u^{\mathrm{corr}}$)} &
        \makebox[\qualw][c]{\fontsize{7}{8}\selectfont +Soft Gating ($G$)} \\[1pt]
        \qualimg{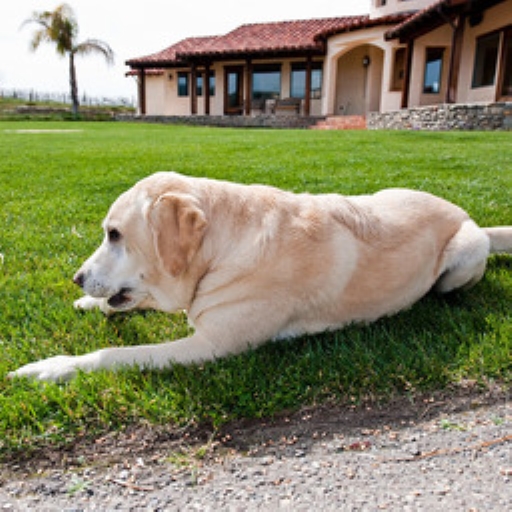} &
        \qualimg{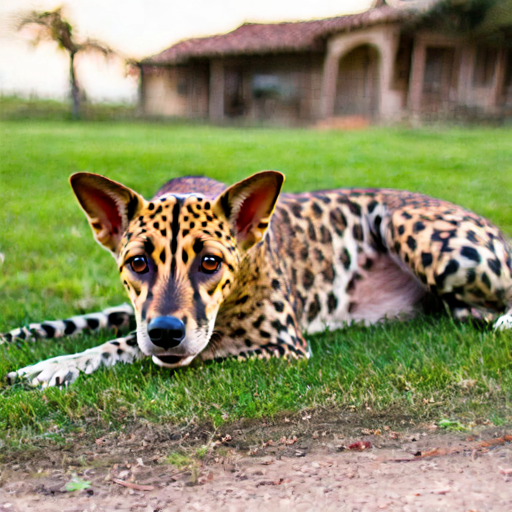} &
        \qualimg{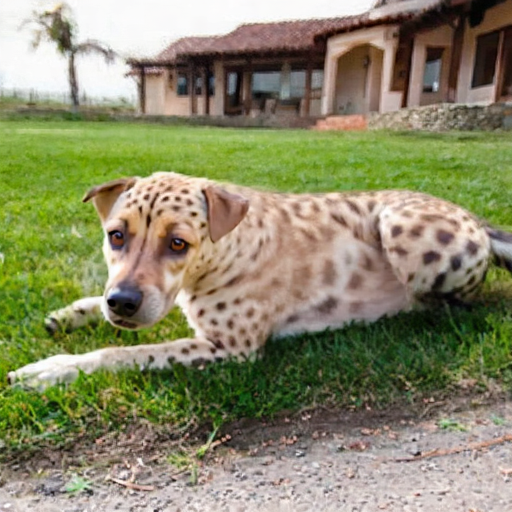} &
        \qualimg{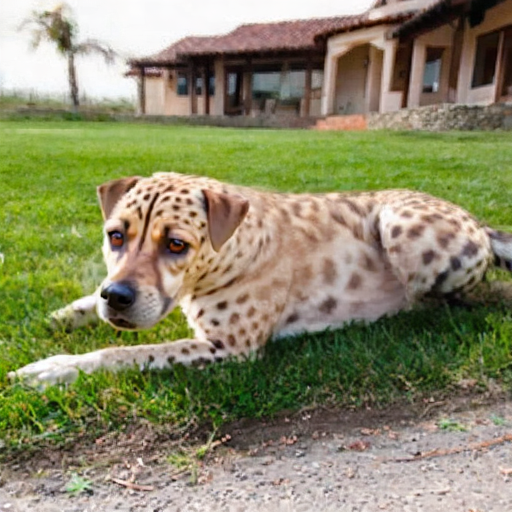} \\
        \multicolumn{4}{c}{\ablationprompt{\textit{white dog} $\rightarrow$ \textcolor{qualpurple}{\textit{leopard dog}}}} \\[1pt]
        \qualimg{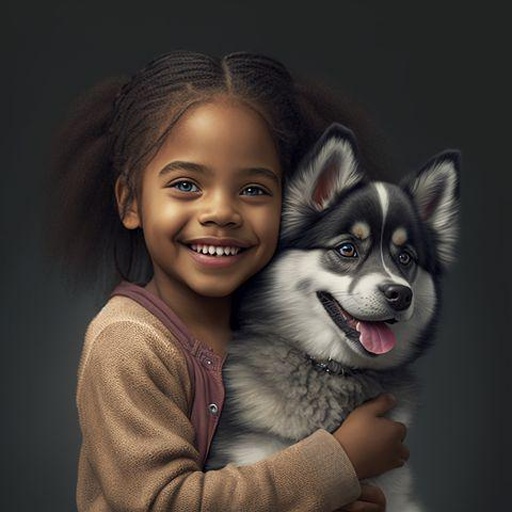} &
        \qualimg{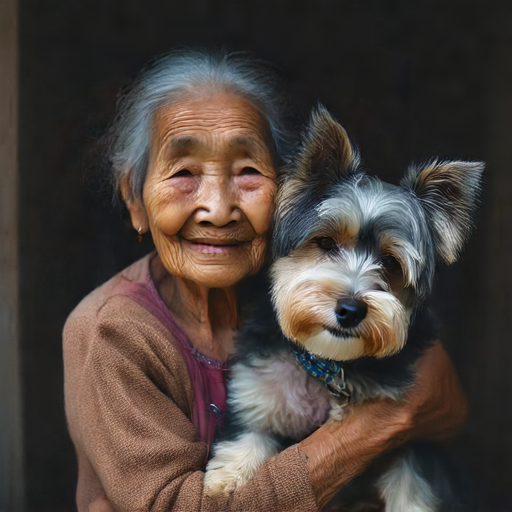} &
        \qualimg{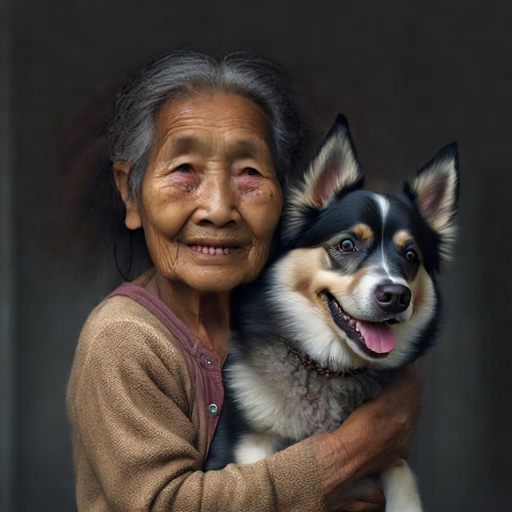} &
        \qualimg{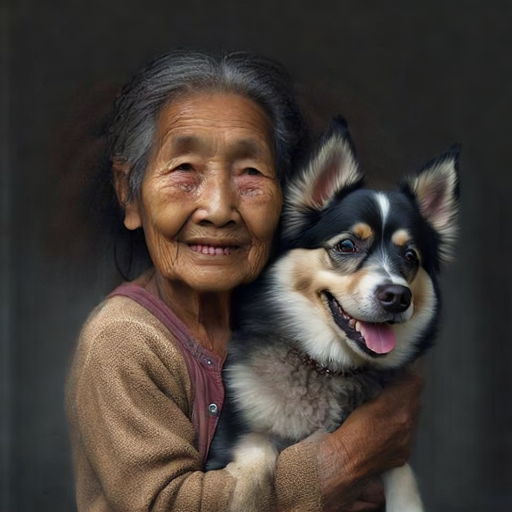} \\
        \multicolumn{4}{c}{\ablationprompt{\textit{girl} $\rightarrow$ \textcolor{qualpurple}{\textit{old woman}}}} \\
    \end{tabular}%
    }
    \sbox0{\ablationgrid}%
    \pgfmathsetmacro{\ablationpromptscale}{\wd0/(0.95*\linewidth)}%
    \def\ablationprompt#1{\scalebox{\ablationpromptscale}{\scalebox{0.8}{\small\sffamily #1}}}
    \resizebox{0.95\linewidth}{!}{\ablationgrid}
    \caption{
        \textbf{Qualitative ablation on FCEdit components.}
        Two qualitative samples; the columns correspond to the first, third, and fourth rows of Table~\ref{tab:ablation_component}.
The Edit Only result (no correction, no gating) fails to preserve the source. Adding correction visibly improves the result, and adding gating makes the edit spatially more precise (\eg, the house in the background becomes clearer).
    }
    \label{fig:ablation_component_qual}
    \par\vspace{4pt}
    \def\FCEditAblationInline{1}
    \input{table/ablation_component}
\end{figure}

\paragraph{Analysis on FCEdit components.}
We analyze the contribution of each FCEdit component through qualitative results (Figure~\ref{fig:ablation_component_qual}) and a quantitative ablation (Table~\ref{tab:ablation_component}).
The edit-only baseline (no correction, no gating) achieves the highest CLIP scores but, as seen in Figure~\ref{fig:ablation_component_qual}, fails to maintain the source identity.
Applying the correction with the rectified-flow SNR schedule, which weights it by timestep only, partially mitigates this drift.
Replacing it with the proposed time--frequency Wiener gain yields the largest improvement, reducing structure distance from 22.71 to 10.88. 
Adding SGD further improves all five preservation metrics while partially recovering both whole-image and edited-region CLIP similarity. The qualitative examples also show that SGD better confines the source-restoring correction to non-edited regions, producing clearer background structures without weakening the intended semantic change. These results demonstrate that FWC provides the primary source-consistency improvement, whereas SGD further refines the spatial balance between source preservation and editability.
 Additional results, analysis, and runtime are reported in \suppref{app:extra}.

%% file: table/quant_user.tex
\begin{table}[t]
\centering
\setlength{\tabcolsep}{0pt}
\renewcommand{\arraystretch}{0.80}
\begin{tabular*}{\columnwidth}{@{\extracolsep{\fill}}lccc@{}}
\toprule
Method & Sem.$\uparrow$ & Cons.$\uparrow$ & Pref.$\uparrow$ \\
\midrule
\textit{FCEdit (Ours)}
& \textbf{37.8} & \textbf{36.8} & \textbf{38.3} \\
\midrule
DRFS~\cite{drfs}
& 18.8 & 18.8 & 17.0 \\
FIA-Edit~\cite{fia_edit}
& 15.3 & \underline{21.2} & 15.5 \\
FlowAlign~\cite{flow_align}
& 13.3 & 14.5 & 10.5 \\
FlowEdit~\cite{flow_edit}
& \underline{24.7} & 16.3 & \underline{20.8} \\
\bottomrule
\end{tabular*}
\caption{\textbf{Human preference study.} Sem., Cons., and Pref. denote
semantic alignment, source consistency, and overall preference. Values are
selection rates (\%). The best and second-best scores are shown in
\textbf{bold} and \underline{underline}, respectively.}
\label{tab:human_pref}
\end{table}

%% file: table/ablation_component.tex
\ifdefined\FCEditAblationInline
    \begingroup
    \captionsetup{type=table}
\else
    \begin{table}[t]
\fi
    \centering
    \setlength{\tabcolsep}{4pt}
    \renewcommand{\arraystretch}{1.10}
    \resizebox{\columnwidth}{!}{%
    \begin{tabular}{@{}cc|ccccc|cc@{}}
        \toprule
        \multicolumn{2}{c|}{Component}
        & \multicolumn{5}{c|}{Preservation}
        & \multicolumn{2}{c}{CLIP} \\
        \cmidrule(r){1-2}\cmidrule(lr){3-7}\cmidrule(l){8-9}
        FWC & SGD
        & SD $\downarrow$ & PSNR $\uparrow$ & LPIPS $\downarrow$
        & MSE $\downarrow$ & SSIM $\uparrow$
        & Whole $\uparrow$ & Edited $\uparrow$ \\
        \midrule
        \xmark & \xmark & 33.89 & 21.19 & 123.09 & 107.90 & 82.20
               & \textbf{28.08} & \textbf{24.40} \\
        time-only & \xmark & 22.71 & 23.27 & 92.21 & 68.34 & 85.56
        & 27.78 & 23.92 \\
        \cmark & \xmark & 10.88 & 27.24 & 48.82 & 30.88 & 89.88
               & 26.93 & 23.19 \\
        \cmark & \cmark & \textbf{10.14} & \textbf{27.35} & \textbf{48.40} & \textbf{29.49} & \textbf{90.03} & 26.95 & 23.26 \\
        \bottomrule
    \end{tabular}%
    }
    \caption{\textbf{Ablation on FCEdit components.} FWC: Frequency-wise Wiener Correction (time- and frequency-wise); SGD: Soft Gate Decomposition. FWC drives the bulk of the preservation gain, and SGD improves all five preservation metrics further.}
    \label{tab:ablation_component}
\ifdefined\FCEditAblationInline
    \endgroup
\else
    \end{table}
\fi

%% file: sec/5_conclusion.tex
\section{Conclusion}
\label{sec:conclusion}
We present FCEdit, an inversion-free editing framework that treats source-consistency correction in rectified-flow models as a reliability-estimation problem. Its frequency-wise Wiener correction derives a closed-form time–frequency gain from the signal and error powers of the endpoint discrepancy. This gain recovers the constant and time-only schedules of prior work as frequency-flat special cases. An energy-guided soft gate further balances editing and correction across space using signals already computed during sampling. As a result, FCEdit substantially improves source preservation while maintaining competitive editability, without external masks or additional network evaluations.

Like other training-free editors, FCEdit relies on the generative prior of its pretrained flow model. Editing pose or shape with large geometric changes may remain challenging. Extending FCEdit to few-step and video editing models is a promising direction for future work.

%% file: main.bib
@String(CVPR= {IEEE Conf. Comput. Vis. Pattern Recog.})

@String(ICCV= {Int. Conf. Comput. Vis.})

@String(ECCV= {Eur. Conf. Comput. Vis.})

@String(NIPS= {Adv. Neural Inform. Process. Syst.})

@String(TIP  = {IEEE Trans. Image Process.})

@String(ICLR = {Int. Conf. Learn. Represent.})

@String(AAAI = {AAAI})

@String(ICML = {Int. Conf. Mach. Learn.})

@String(CVPR  = {CVPR})

@String(ICCV  = {ICCV})

@String(ECCV  = {ECCV})

@String(NIPS  = {NeurIPS})

@String(TIP   = {IEEE TIP})

@String(ICLR  = {ICLR})

@String(ICML = {ICML})

@misc{prolific2026,
  author       = {{Prolific}},
  title        = {{Prolific}: Online participant recruitment platform},
  year         = {2026},
  howpublished = {\url{https://www.prolific.com}},
  note         = {Accessed: 2026-07-29}
}

@article{Schaaf1996ModellingTP,
  author  = {van der Schaaf, A. and van Hateren, J. H.},
  title   = {Modelling the power spectra of natural images:
             Statistics and information},
  journal = {Vision Research},
  volume  = {36},
  number  = {17},
  pages   = {2759--2770},
  year    = {1996}
}

@article{Goldstein1998AMR,
  author  = {Goldstein, J. Scott and Reed, Irving S. and Scharf, Louis L.},
  title   = {A multistage representation of the {Wiener} filter
             based on orthogonal projections},
  journal = {IEEE Trans. Inf. Theory},
  volume  = {44},
  number  = {7},
  pages   = {2943--2959},
  year    = {1998},
  doi     = {10.1109/18.737524}
}

@book{Haykin1986AdaptiveFT,
  author    = {Haykin, Simon},
  title     = {Adaptive Filter Theory},
  publisher = {Prentice-Hall},
  year      = {1986}
}

@book{wiener1949extrapolation,
  author    = {Wiener, Norbert},
  title     = {Extrapolation, Interpolation, and Smoothing of
               Stationary Time Series: With Engineering Applications},
  publisher = {MIT Press},
  year      = {1949}
}

@article{Simoncelli2001NaturalIS,
  author  = {Simoncelli, Eero P. and Olshausen, Bruno A.},
  title   = {Natural image statistics and neural representation},
  journal = {Annu. Rev. Neurosci.},
  volume  = {24},
  pages   = {1193--1216},
  year    = {2001}
}

@article{field1987relations,
  author  = {Field, David J.},
  title   = {Relations between the statistics of natural images
             and the response properties of cortical cells},
  journal = {J. Opt. Soc. Am. A},
  volume  = {4},
  number  = {12},
  pages   = {2379--2394},
  year    = {1987}
}

@article{Mandelbrot1967SomeNW,
  author  = {Mandelbrot, Benoit B.},
  title   = {Some noises with {$1/f$} spectrum, a bridge between
             direct current and white noise},
  journal = {IEEE Trans. Inf. Theory},
  volume  = {13},
  pages   = {289--298},
  year    = {1967}
}

@inproceedings{songdenoising,
  author    = {Song, Jiaming and Meng, Chenlin and Ermon, Stefano},
  title     = {Denoising diffusion implicit models},
  booktitle = ICLR,
  year      = {2021}
}

@article{Ho2022ClassifierFreeDG,
  author  = {Ho, Jonathan and Salimans, Tim},
  title   = {Classifier-free diffusion guidance},
  journal = {arXiv preprint arXiv:2207.12598},
  year    = {2022}
}

@inproceedings{lipman2023flow,
  author    = {Lipman, Yaron and Chen, Ricky T. Q. and
               Ben-Hamu, Heli and Nickel, Maximilian and Le, Matt},
  title     = {Flow matching for generative modeling},
  booktitle = ICLR,
  year      = {2023}
}

@inproceedings{liuflow,
  author    = {Liu, Xingchao and Gong, Chengyue and Liu, Qiang},
  title     = {Flow straight and fast: Learning to generate and
               transfer data with rectified flow},
  booktitle = ICLR,
  year      = {2023}
}

@inproceedings{rissanen2023generative,
  author    = {Rissanen, Severi and Heinonen, Markus and Solin, Arno},
  title     = {Generative modelling with inverse heat dissipation},
  booktitle = ICLR,
  year      = {2023}
}

@inproceedings{Hoogeboom2022BlurringDM,
  author    = {Hoogeboom, Emiel and Salimans, Tim},
  title     = {Blurring diffusion models},
  booktitle = ICLR,
  year      = {2023}
}

@inproceedings{hertzprompt,
  author    = {Hertz, Amir and Mokady, Ron and Tenenbaum, Jay and
               Aberman, Kfir and Pritch, Yael and Cohen-Or, Daniel},
  title     = {{Prompt-to-Prompt} image editing with
               cross-attention control},
  booktitle = ICLR,
  year      = {2023}
}

@inproceedings{null_text_inversion,
  author    = {Mokady, Ron and Hertz, Amir and Aberman, Kfir and
               Pritch, Yael and Cohen-Or, Daniel},
  title     = {Null-text inversion for editing real images
               using guided diffusion models},
  booktitle = CVPR,
  pages     = {6038--6047},
  year      = {2023}
}

@inproceedings{blended_diffusion,
  author    = {Avrahami, Omri and Lischinski, Dani and Fried, Ohad},
  title     = {Blended diffusion for text-driven editing
               of natural images},
  booktitle = CVPR,
  pages     = {18208--18218},
  year      = {2022}
}

@inproceedings{su2022dual,
  author    = {Su, Xuan and Song, Jiaming and Meng, Chenlin and
               Ermon, Stefano},
  title     = {Dual diffusion implicit bridges for
               image-to-image translation},
  booktitle = ICLR,
  year      = {2023}
}

@inproceedings{pix2pix_zero,
  author    = {Parmar, Gaurav and Singh, Krishna Kumar and
               Zhang, Richard and Li, Yijun and Lu, Jingwan and
               Zhu, Jun-Yan},
  title     = {Zero-shot image-to-image translation},
  booktitle = {ACM SIGGRAPH Conference Proceedings},
  pages     = {1--11},
  year      = {2023}
}

@inproceedings{plug_and_play,
  author    = {Tumanyan, Narek and Geyer, Michal and Bagon, Shai and
               Dekel, Tali},
  title     = {Plug-and-play diffusion features for text-driven
               image-to-image translation},
  booktitle = CVPR,
  pages     = {1921--1930},
  year      = {2023}
}

@inproceedings{masactrl,
  author    = {Cao, Mingdeng and Wang, Xintao and Qi, Zhongang and
               Shan, Ying and Qie, Xiaohu and Zheng, Yinqiang},
  title     = {{MasaCtrl}: Tuning-free mutual self-attention control
               for consistent image synthesis and editing},
  booktitle = ICCV,
  pages     = {22560--22570},
  year      = {2023}
}

@inproceedings{mengsdedit,
  author    = {Meng, Chenlin and He, Yutong and Song, Yang and
               Song, Jiaming and Wu, Jiajun and Zhu, Jun-Yan and
               Ermon, Stefano},
  title     = {{SDEdit}: Guided image synthesis and editing
               with stochastic differential equations},
  booktitle = ICLR,
  year      = {2022}
}

@inproceedings{diff_edit,
  author    = {Couairon, Guillaume and Verbeek, Jakob and
               Schwenk, Holger and Cord, Matthieu},
  title     = {{DiffEdit}: Diffusion-based semantic image editing
               with mask guidance},
  booktitle = ICLR,
  year      = {2023}
}

@inproceedings{dengfireflow,
  author    = {Deng, Yingying and He, Xiangyu and Mei, Changwang and
               Wang, Peisong and Tang, Fan},
  title     = {{FireFlow}: Fast inversion of rectified flow
               for image semantic editing},
  booktitle = ICML,
  pages     = {13110--13128},
  year      = {2025}
}

@inproceedings{wang2025taming,
  author    = {Wang, Jiangshan and Pu, Junfu and Qi, Zhongang and
               Guo, Jiayi and Ma, Yue and Huang, Nisha and
               Chen, Yuxin and Li, Xiu and Shan, Ying},
  title     = {Taming rectified flow for inversion and editing},
  booktitle = ICML,
  pages     = {64044--64058},
  year      = {2025}
}

@inproceedings{Rout2024SemanticII,
  author    = {Rout, Litu and Chen, Yujia and Ruiz, Nataniel and
               Caramanis, Constantine and Shakkottai, Sanjay and
               Chu, Wen-Sheng},
  title     = {Semantic image inversion and editing using
               rectified stochastic differential equations},
  booktitle = ICLR,
  year      = {2025}
}

@inproceedings{kv_edit,
  author    = {Zhu, Tianrui and Zhang, Shiyi and Shao, Jiawei and
               Tang, Yansong},
  title     = {{KV-Edit}: Training-free image editing
               for precise background preservation},
  booktitle = ICCV,
  pages     = {16607--16617},
  year      = {2025}
}

@inproceedings{flow_edit,
  author    = {Kulikov, Vladimir and Kleiner, Matan and
               Huberman-Spiegelglas, Inbar and Michaeli, Tomer},
  title     = {{FlowEdit}: Inversion-free text-based editing
               using pre-trained flow models},
  booktitle = ICCV,
  pages     = {19721--19730},
  year      = {2025}
}

@inproceedings{ft_edit,
  author    = {Xu, Pengcheng and Jiang, Boyuan and Hu, Xiaobin and
               Luo, Donghao and He, Qingdong and Zhang, Jiangning and
               Wang, Chengjie and Wu, Yunsheng and Ling, Charles and
               Wang, Boyu},
  title     = {Unveil inversion and invariance in flow transformer
               for versatile image editing},
  booktitle = CVPR,
  pages     = {28479--28489},
  year      = {2025}
}

@inproceedings{flow_align,
  author    = {Kim, Jeongsol and Hong, Yeobin and Park, Jonghyun and
               Ye, Jong Chul},
  title     = {{FlowAlign}: Trajectory-regularized, inversion-free
               flow-based image editing},
  booktitle = ICLR,
  year      = {2026}
}

@inproceedings{xiednaedit,
  author    = {Xie, Chenxi and Li, Minghan and Li, Shuai and
               Wu, Yuhui and Yi, Qiaosi and Zhang, Lei},
  title     = {{DNAEdit}: Direct noise alignment for
               text-guided rectified flow editing},
  booktitle = NIPS,
  volume    = {38},
  year      = {2025}
}

@inproceedings{fia_edit,
  author    = {Yang, Kaixiang and Shen, Boyang and Li, Xin and
               Dai, Yuchen and Luo, Yuxuan and Ma, Yueran and
               Fang, Wei and Li, Qiang and Wang, Zhiwei},
  title     = {{FIA-Edit}: Frequency-interactive attention
               for efficient and high-fidelity inversion-free
               text-guided image editing},
  booktitle = AAAI,
  volume    = {40},
  number    = {14},
  pages     = {11613--11621},
  year      = {2026},
  doi       = {10.1609/aaai.v40i14.38145}
}

@inproceedings{drfs,
  author    = {Beaudouin, Gaspard and Li, Minghan and Kim, Jaeyeon and
               Yoon, Sung-Hoon and Wang, Mengyu},
  title     = {Delta rectified flow sampling for text-to-image editing},
  booktitle = CVPR,
  pages     = {18662--18672},
  year      = {2026}
}

@inproceedings{chord_edit,
  author    = {Lu, Liangsi and Chen, Xuhang and Guo, Minzhe and
               Li, Shichu and Wang, Jingchao and Shi, Yang},
  title     = {{ChordEdit}: One-step low-energy transport
               for image editing},
  booktitle = CVPR,
  pages     = {14398--14407},
  year      = {2026}
}

@inproceedings{split_flow,
  author    = {Yoon, Sung-Hoon and Li, Minghan and Beaudouin, Gaspard and
               Wen, Congcong and Azhar, Muhammad Rafay and Wang, Mengyu},
  title     = {{SplitFlow}: Flow decomposition for
               inversion-free text-to-image editing},
  booktitle = NIPS,
  volume    = {38},
  year      = {2025}
}

@inproceedings{yangdirectedit,
  author    = {Yang, Desong and Ye, Mang},
  title     = {{DirectEdit}: Step-level accurate inversion
               for flow-based image editing},
  booktitle = ICML,
  pages     = {146750--146765},
  year      = {2026}
}

@inproceedings{exploring_mmdit,
  author    = {Shin, Joonghyuk and Hwang, Alchan and Kim, Yujin and
               Kim, Daneul and Park, Jaesik},
  title     = {Exploring multimodal diffusion transformers
               for enhanced prompt-based image editing},
  booktitle = ICCV,
  year      = {2025}
}

@inproceedings{SD3,
  author    = {Esser, Patrick and Kulal, Sumith and Blattmann, Andreas and
               Entezari, Rahim and M{\"u}ller, Jonas and Saini, Harry and
               Levi, Yam and Lorenz, Dominik and Sauer, Axel and
               Boesel, Frederic and others},
  title     = {Scaling rectified flow transformers
               for high-resolution image synthesis},
  booktitle = ICML,
  year      = {2024}
}

@misc{SD35_medium,
  author       = {{Stability AI}},
  title        = {{Stable Diffusion 3.5 Medium}},
  year         = {2024},
  howpublished = {\url{https://huggingface.co/stabilityai/stable-diffusion-3.5-medium}},
  note         = {Accessed: 2024-11-14}
}

@misc{flux,
  author       = {{Black Forest Labs}},
  title        = {{FLUX}: Official code repository},
  year         = {2024},
  howpublished = {\url{https://github.com/black-forest-labs/flux}},
  note         = {Accessed: 2024-11-14}
}

@misc{flux1_dev,
  author       = {{Black Forest Labs}},
  title        = {{FLUX.1-dev}: Official model weights},
  year         = {2024},
  howpublished = {\url{https://huggingface.co/black-forest-labs/FLUX.1-dev}},
  note         = {Accessed: 2024-11-14}
}

@inproceedings{pie_bench,
  author    = {Ju, Xuan and Zeng, Ailing and Bian, Yuxuan and
               Liu, Shaoteng and Xu, Qiang},
  title     = {{PnP} inversion: Boosting diffusion-based editing
               with 3 lines of code},
  booktitle = ICLR,
  year      = {2024}
}

@inproceedings{Radford2021LearningTV,
  author    = {Radford, Alec and Kim, Jong Wook and Hallacy, Chris and
               Ramesh, Aditya and Goh, Gabriel and Agarwal, Sandhini and
               Sastry, Girish and Askell, Amanda and Mishkin, Pamela and
               Clark, Jack and Krueger, Gretchen and Sutskever, Ilya},
  title     = {Learning transferable visual models
               from natural language supervision},
  booktitle = ICML,
  pages     = {8748--8763},
  year      = {2021}
}

@article{Wang2004ImageQA,
  author  = {Wang, Zhou and Bovik, Alan Conrad and Sheikh, Hamid R. and
             Simoncelli, Eero P.},
  title   = {Image quality assessment: From error visibility
             to structural similarity},
  journal = TIP,
  volume  = {13},
  number  = {4},
  pages   = {600--612},
  year    = {2004}
}

@inproceedings{Zhang2018TheUE,
  author    = {Zhang, Richard and Isola, Phillip and Efros, Alexei A. and
               Shechtman, Eli and Wang, Oliver},
  title     = {The unreasonable effectiveness of deep features
               as a perceptual metric},
  booktitle = CVPR,
  pages     = {586--595},
  year      = {2018}
}

@inproceedings{xu2024infedit,
  author    = {Xu, Sihan and Huang, Yidong and Pan, Jiayi and
               Ma, Ziqiao and Chai, Joyce},
  title     = {Inversion-free image editing with language-guided
               diffusion models},
  booktitle = CVPR,
  pages     = {9452--9461},
  year      = {2024}
}

@inproceedings{avrahami2025stableflow,
  author    = {Avrahami, Omri and Patashnik, Or and Fried, Ohad and
               Nemchinov, Egor and Aberman, Kfir and Lischinski, Dani and
               Cohen-Or, Daniel},
  title     = {Stable {Flow}: Vital layers for training-free image editing},
  booktitle = CVPR,
  pages     = {7877--7888},
  year      = {2025}
}

@inproceedings{jiao2026unieditflow,
  author    = {Jiao, Guanlong and Huang, Biqing and Wang, Kuan-Chieh and
               Liao, Renjie},
  title     = {{UniEdit-Flow}: Unleashing inversion and editing
               in the era of flow models},
  booktitle = ICLR,
  year      = {2026}
}

@article{bfl2025kontext,
  author    = {{Black Forest Labs} and Batifol, Stephen and
               Blattmann, Andreas and Boesel, Frederic and others},
  title     = {{FLUX.1 Kontext}: Flow matching for in-context image
               generation and editing in latent space},
  journal   = {arXiv preprint arXiv:2506.15742},
  year      = {2025}
}

@inproceedings{wu2024turboedit,
  author    = {Wu, Zongze and Kolkin, Nicholas and Brandt, Jonathan and
               Zhang, Richard and Shechtman, Eli},
  title     = {{TurboEdit}: Instant text-based image editing},
  booktitle = ECCV,
  pages     = {365--381},
  year      = {2024}
}

@inproceedings{nguyen2025swiftedit,
  author    = {Nguyen, Trong-Tung and Nguyen, Quang and Nguyen, Khoi and
               Tran, Anh and Pham, Cuong},
  title     = {{SwiftEdit}: Lightning fast text-guided image editing
               via one-step diffusion},
  booktitle = CVPR,
  pages     = {21492--21501},
  year      = {2025}
}

@inproceedings{brooks2023instructpix2pix,
  author    = {Brooks, Tim and Holynski, Aleksander and Efros, Alexei A.},
  title     = {{InstructPix2Pix}: Learning to follow image editing
               instructions},
  booktitle = CVPR,
  pages     = {18392--18402},
  year      = {2023}
}

@inproceedings{wallace2023edict,
  author    = {Wallace, Bram and Gokul, Akash and Naik, Nikhil},
  title     = {{EDICT}: Exact diffusion inversion via coupled
               transformations},
  booktitle = CVPR,
  pages     = {22532--22541},
  year      = {2023}
}

@inproceedings{huberman2024editfriendly,
  author    = {Huberman-Spiegelglas, Inbar and Kulikov, Vladimir and
               Michaeli, Tomer},
  title     = {An edit friendly {DDPM} noise space: Inversion and
               manipulations},
  booktitle = CVPR,
  pages     = {12469--12478},
  year      = {2024}
}

@inproceedings{ho2020ddpm,
  author    = {Ho, Jonathan and Jain, Ajay and Abbeel, Pieter},
  title     = {Denoising diffusion probabilistic models},
  booktitle = NIPS,
  volume    = {33},
  pages     = {6840--6851},
  year      = {2020}
}

@inproceedings{rombach2022ldm,
  author    = {Rombach, Robin and Blattmann, Andreas and Lorenz, Dominik and
               Esser, Patrick and Ommer, Bj{\"o}rn},
  title     = {High-resolution image synthesis with latent diffusion models},
  booktitle = CVPR,
  pages     = {10684--10695},
  year      = {2022}
}

@inproceedings{peebles2023dit,
  author    = {Peebles, William and Xie, Saining},
  title     = {Scalable diffusion models with transformers},
  booktitle = ICCV,
  pages     = {4195--4205},
  year      = {2023}
}

@inproceedings{tumanyan2022splicing,
  author    = {Tumanyan, Narek and Bar-Tal, Omer and Bagon, Shai and
               Dekel, Tali},
  title     = {Splicing {ViT} features for semantic appearance transfer},
  booktitle = CVPR,
  pages     = {10748--10757},
  year      = {2022}
}

@inproceedings{caron2021dino,
  author    = {Caron, Mathilde and Touvron, Hugo and Misra, Ishan and
               J{\'e}gou, Herv{\'e} and Mairal, Julien and
               Bojanowski, Piotr and Joulin, Armand},
  title     = {Emerging properties in self-supervised vision transformers},
  booktitle = ICCV,
  pages     = {9650--9660},
  year      = {2021}
}

@article{falck2025fourier,
  author    = {Falck, Fabian and Pandeva, Teodora and Zahirnia, Kiarash and
               Lawrence, Rachel and Turner, Richard and Meeds, Edward and
               Zazo, Javier and Karmalkar, Sushrut},
  title     = {A {Fourier} space perspective on diffusion models},
  journal   = {arXiv preprint arXiv:2505.11278},
  year      = {2025}
}

@article{torralba2003statistics,
  author    = {Torralba, Antonio and Oliva, Aude},
  title     = {Statistics of natural image categories},
  journal   = {Network: Computation in Neural Systems},
  volume    = {14},
  number    = {3},
  pages     = {391--412},
  year      = {2003}
}

@inproceedings{hertz2023dds,
  author    = {Hertz, Amir and Aberman, Kfir and Cohen-Or, Daniel},
  title     = {Delta denoising score},
  booktitle = ICCV,
  pages     = {2328--2337},
  year      = {2023}
}

@inproceedings{ren2025fds,
  author    = {Ren, Yufan and Jiang, Zicong and Zhang, Tong and
               Forchhammer, S{\o}ren and S{\"u}sstrunk, Sabine},
  title     = {{FDS}: Frequency-aware denoising score for text-guided
               latent diffusion image editing},
  booktitle = CVPR,
  pages     = {2651--2660},
  year      = {2025}
}
